\PassOptionsToPackage{table}{xcolor}
\documentclass{article}

\usepackage{iftex}
\ifPDFTeX
  \usepackage{iclr2027_conference,times}
\else
  \usepackage{iclr2027_conference}
  \usepackage{fontspec}
\fi

\usepackage{amsmath,amssymb,mathtools}
\usepackage{booktabs,tabularx,array,multirow,makecell}
\usepackage{xcolor}
\usepackage{graphicx}
\usepackage{microtype}
\usepackage{enumitem}
\usepackage{etoc} %
\newsavebox{\AppendixContentsBox}
\usepackage{placeins}
\usepackage{float}
\usepackage{needspace}
\usepackage{url}
\usepackage{hyperref}
\usepackage{cleveref}
\crefname{figure}{Figure}{Figures}
\Crefname{figure}{Figure}{Figures}
\usepackage{listings}
\usepackage{tcolorbox}
\tcbuselibrary{breakable,skins,listings}

\renewenvironment{abstract}
  {\par
   \centerline{\large\scshape Abstract}
   \vspace{0.5ex}
   \begin{list}{}{
     \setlength{\leftmargin}{0.03\textwidth}
     \setlength{\rightmargin}{0.03\textwidth}
     \setlength{\topsep}{0pt}
     \setlength{\partopsep}{2pt}
   }\item\relax}
  {\end{list}\par}
  
\usepackage[labelfont=bf,font=small]{caption}

\definecolor{Ink}{HTML}{2D3540}
\definecolor{Primary}{HTML}{405A70}
\definecolor{Accent}{HTML}{5D7774}
\definecolor{CardinalRed}{HTML}{405A70}
\definecolor{CardinalDark}{HTML}{334B60}
\definecolor{DeepTeal}{HTML}{526C69}
\definecolor{TableHeader}{HTML}{E9EEF1}
\definecolor{TableHeaderStrong}{HTML}{DEE6EA}
\definecolor{TableStripe}{HTML}{F7F8F9}
\definecolor{RuleGray}{HTML}{B9C2C8}
\definecolor{LinkBlue}{HTML}{245B8A}
\definecolor{TakeawayFill}{HTML}{F2F6F5}

\newcolumntype{L}[1]{>{\raggedright\arraybackslash}p{#1}}
\newcolumntype{Y}{>{\raggedright\arraybackslash}X}

\newcommand{\Full}{\textsc{Full}}
\newcommand{\Coarse}{\textsc{Coarse}}
\newcommand{\Least}{\textsc{Least}}

\definecolor{ModelMini}{HTML}{3E668C}
\definecolor{ModelGPT}{HTML}{B6683C}
\definecolor{ModelGemini}{HTML}{8C5669}
\definecolor{ModelSonnet}{HTML}{756395}
\definecolor{FindingFill}{HTML}{F0F7FA}
\newcounter{finding}
\newenvironment{finding}[1]{%
  \begin{tcolorbox}[
    enhanced,
    colback=FindingFill,
    colframe=Ink,
    boxrule=0.4pt,
    arc=2pt,
    boxsep=0pt,
    left=6pt,right=6pt,top=4pt,bottom=4pt,
    before skip=6pt,after skip=6pt,
    fontupper=\normalsize\normalfont
  ]%
  \refstepcounter{finding}%
  \noindent\textbf{Finding \thefinding. #1}
}{\end{tcolorbox}}

\newtcolorbox{takeaway}[1][]{
  enhanced,
  breakable,
  colback=TakeawayFill,
  frame hidden,
  borderline west={1.2pt}{0pt}{Accent},
  coltitle=CardinalDark,
  fonttitle=\small\bfseries,
  fontupper=\small,
  title={#1},
  sharp corners,
  left=7pt,right=6pt,top=5pt,bottom=5pt,
  before skip=8pt,after skip=8pt
}

\lstdefinestyle{prompt}{
  basicstyle=\ttfamily\fontsize{7.5}{8.3}\selectfont,
  columns=fullflexible,
  breaklines=true,
  breakatwhitespace=true,
  keepspaces=true,
  frame=l,
  framerule=1.2pt,
  rulecolor=\color{Primary},
  backgroundcolor=\color{TableStripe},
  xleftmargin=9pt,
  xrightmargin=6pt,
  framexleftmargin=7pt,
  framexrightmargin=5pt,
  framesep=5pt,
  aboveskip=5pt,
  belowskip=5pt,
  abovecaptionskip=3pt,
  belowcaptionskip=3pt,
  title=\textcolor{Primary}{\small\bfseries Prompt template},
  emph={OBJECTIVE,GAME,RULES,PUBLIC,INVENTORIES,PRIVATE,INFORMATION,STATE,HISTORY,YOUR,ROLE,AND,BACKGROUND,UNCERTAINTY,AVAILABLE,TO,YOU,CURRENT,DIALOGUE,SO,FAR,LEGAL,ACTIONS},
  emphstyle=\color{Primary}\bfseries,
  showstringspaces=false
}

\definecolor{PromptSkyBlue}{HTML}{5BA9D0}
\newtcblisting{promptbox}[2][]{
  enhanced,
  breakable,
  listing only,
  listing options={style=prompt,frame=none,backgroundcolor={},
 xleftmargin=0pt,xrightmargin=0pt,aboveskip=0pt,belowskip=0pt,title={},#1},
  colback=TableStripe,
  colframe=Primary,
  colbacktitle=PromptSkyBlue,
  coltitle=white,
  fonttitle=\small\bfseries,
  title={#2},
  title after break={#2\ (continued)},
  boxrule=0.6pt,
  arc=2pt,
  boxsep=0pt,
  left=8pt,right=8pt,top=6pt,bottom=6pt,
  toptitle=4pt,bottomtitle=4pt,
  before skip=8pt,after skip=8pt
}


\hypersetup{
  colorlinks=true,
  linkcolor=LinkBlue,
  citecolor=LinkBlue,
  urlcolor=LinkBlue,
  filecolor=LinkBlue,
  pdftitle={Empirical Information Design for Multi-Agent LLM Systems},
  pdfauthor={Mete Erdogan, Mirac Suzgun, Aneesh Pappu, Guowei Xu,
Batu El, Andrew Adams Schoen, Furong Huang, John P Dickerson,
Mert Pilanci, James Zou}
}

\title{Who Should Know What? Empirical Information Design for Multi-Agent LLM Economies}

\author{
\textbf{Mete Erdogan\textsuperscript{1}, \;
Mirac Suzgun\textsuperscript{1}, \;
Aneesh Pappu\textsuperscript{1}, \;
Guowei Xu\textsuperscript{1}, \;
Batu El\textsuperscript{1}} \\
\textbf{Andrew A. Schoen\textsuperscript{3}, \;
Furong Huang\textsuperscript{2,3}, \;
John P. Dickerson\textsuperscript{2,3}, \;
Mert Pilanci\textsuperscript{1}, \;
James Zou\textsuperscript{1}}
\\[0.5em]
{\normalfont\small
\textsuperscript{1}Stanford University \quad
\textsuperscript{2}University of Maryland, College Park \quad
\textsuperscript{3}All Purpose AI\hspace{0.6cm}}
}

\iclrfinalcopy

\begin{document}
\etocdepthtag.toc{main}
\raggedbottom
\maketitle
\lhead{}

\begin{abstract}

The performance of multi-agent LLM systems depends not only on the capabilities of the underlying models but also on information-sharing policies, which govern what information each agent observes and at what level of detail. But does providing AI agents with more detailed information always improve collective outcomes, or can transparency paradoxically degrade efficiency? To investigate this, we introduce \emph{empirical information design}, an experimental framework that systematically compares disclosure policies across economic markets populated by LLM agents, while holding interaction rules and models fixed. We evaluate ten frontier LLMs in multi-resource trade, bilateral bargaining, and double auctions, using ground-truth utilities to decouple identified opportunities from realized outcomes. We find that finer information disclosure yields highly model-dependent results. In trade, revealing exact valuations rather than ordinal preference rankings reduces realized gains for GPT-6 Sol (from $98\%$ to $91\%$ of maximum) yet increases them for Claude Sonnet 5 (from $72\%$ to $86\%$). In bargaining, exact information leads Grok 4.7 to produce less balanced agreements, whereas Claude Opus 5.5 divides gains symmetrically. In double auctions, full disclosure inflates price demands, triggering market breakdown and lowering welfare for Gemini 3.1 Pro ($97\%$ to $78\%$), Claude Opus 5.5 ($94\%$ to $64\%$), and Grok 4.7 ($97\%$ to $62\%$). Overall, these results suggest that model selection and information design must be co-optimized, as LLMs vary in how their strategic responses to information impact system objectives. {\renewcommand{\thefootnote}{\fnsymbol{footnote}}%
\footnote[4]{{\fontsize{8.8}{2}\selectfont
We provide our code and data at:
\href{https://github.com/meterdogan07/agent-information-design}{\texttt{github.com/meterdogan07/agent-information-design}}}}}\end{abstract}

{\renewcommand{\thefootnote}{\fnsymbol{footnote}}%
\footnotetext[0]{\fontsize{8.8}{2}Correspondence to: \texttt{\{merdogan, msuzgun, jamesz\}@stanford.edu}}

\begin{figure}[!h]
  \centering
  \includegraphics[width=\textwidth]{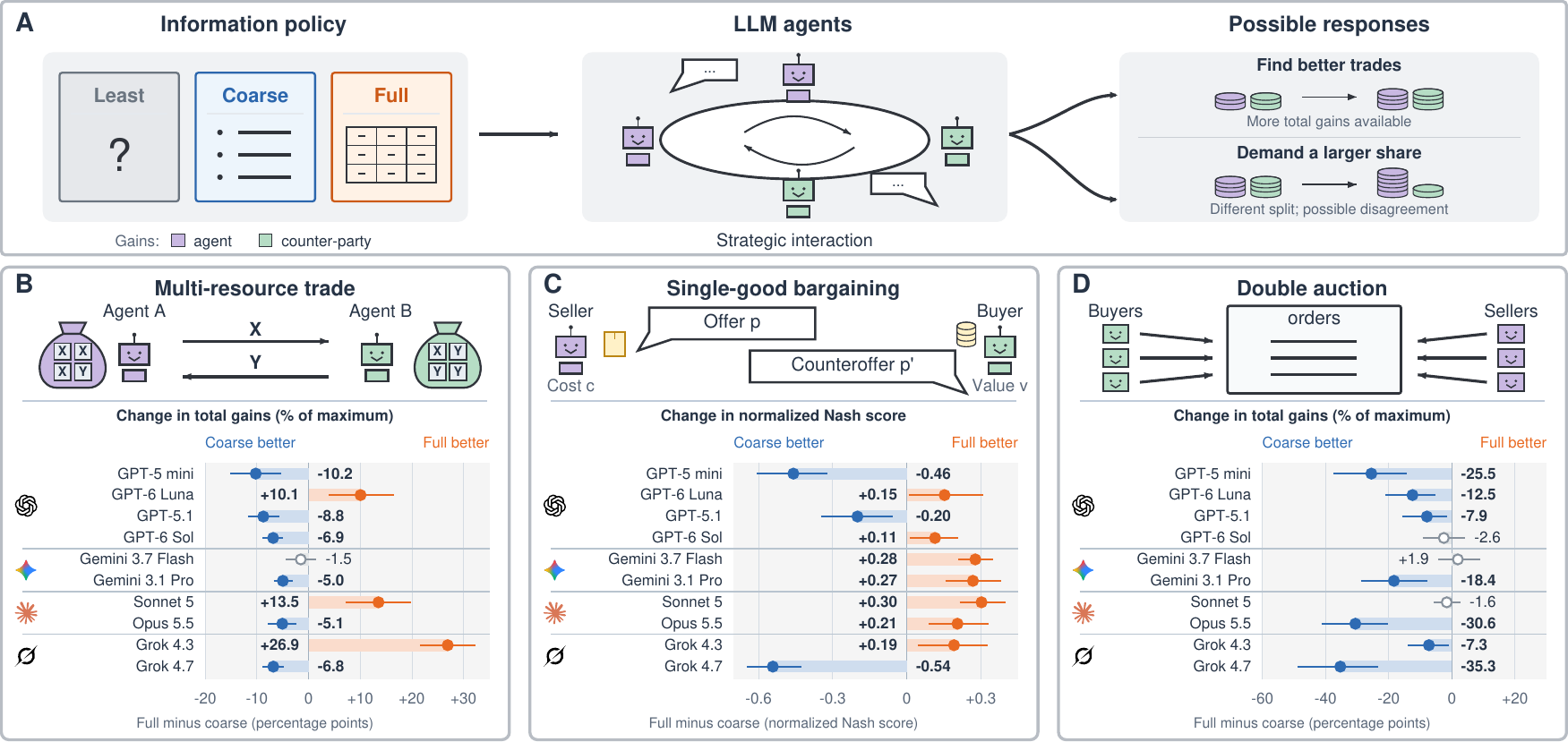} %
  {\small \caption{\textbf{(A) Information policy and agent responses.} Disclosure shapes how agents interact and the outcomes they achieve. \textbf{(B--D) Economic settings and results.} Each panel combines a setting with its disclosure effects: multi-resource trade (B), single-good bargaining (C), and double auctions (D). Bars show how realized performance changes from coarse to full disclosure, with increases favoring full, and decreases favoring coarse information. Trade and auction results are expressed as percentages of maximum available gains; and bargaining shows changes in the normalized Nash score, assigning zero to disagreements. \looseness=-2}
  \label{fig:architecture}}
\end{figure}

\clearpage

\section{Introduction}
\label{sec:introduction}

In December $2025$, Anthropic ran an unusual, futuristic marketplace experiment for a week. About seventy people gave Claude AI agents\footnote{We use the terms ``AI'' and ``LLM'' interchangeably, noting that AI agents need not be language-based.} authority to bargain on their behalf. In this study, called Project Deal, Anthropic employees informed their AI agents about what they wanted to buy or sell, what prices they hoped to receive or were willing to pay, and how they wanted to negotiate. The humans then stepped aside. And their agents posted listings, made offers and counteroffers, and closed $186$ deals involving more than \$$4{,}000$. At the end of the experiments, the participants brought in and exchanged the physical goods and completed the exchanges their agents had negotiated~\citep{anthropic2026deal}. \looseness=-2

Delegating bargaining power to an AI agent creates a critical information-design problem. To represent the wishes and constraints of its principal well, an agent may need, among other things, private information that could fundamentally change the strategy of its counter-party (the opposing agent) if disclosed. A buyer's agent, for instance, may need to know that its user wishes to pay \$$500$ for a product but would be happy to pay as much as \$$1{,}000$ if absolutely necessary. That ceiling helps inform the buyer's agent to decide when to accept, counter, and walk away entirely. To the seller's agent, however, the same figure reveals how much room there is to demand a higher price before losing the sale altogether. Hence, information an agent needs to represent its principal is \emph{not} necessarily the same information that should be revealed to its counter-party.

This distinction motivates an information-design protocol for multi-agent systems. A market designer must decide \emph{not only} which models act, what tools they use, and how they interact, \emph{but also} what information each agent receives. A counter-party might observe an exact reservation price, a coarser signal such as preference ranking or the existence of a mutually beneficial agreement, or no additional information at all. Or exact information might be provided to a mediator who uses it to construct an agreement without revealing any preferential details to the negotiating agents. Overall, we refer to these choices as \emph{information policies}---what is disclosed, to whom, and with what level of detail and precision. Such information policies can materially change what agents infer, what trades they offer, what terms they demand, what compromises they make, and the agreements they eventually reach. 

In an economic exchange, information can support \emph{value discovery}, by helping agents identify transactions that create gains for the participants involved. But the same information can also support what is called \emph{value capture}, by helping an agent decide how much of those gains it can secure for itself. Exact values could therefore help agents identify a better trade while inducing demands that shift gains toward one side or prevent agreement altogether. Additional information can also improve the existing opportunities that agents identify without improving the outcomes they realize.

Here, we study this problem through the lens of \emph{empirical information design}. We conduct controlled experiments that vary the underlying information policy while holding markets, models, incentives, instructions, and interaction rules fixed. We study multi-resource trade, single-good bargaining, and double auctions, in which agents pursue their own gains, and ground-truth utilities enable us to measure both the opportunities they identify and the outcomes they realize. Across these empirical settings, we systematically vary both the  precision and recipient of private information and examine how disclosure changes proposals, demands, agreements, and distribution of gains and overall welfare. In other words, we ask:  \emph{When does more information help AI agents create more value, and when does it lead them to demand terms that worsen the system's outcome in an economic exchange?}

These questions matter more as AI agents act on behalf of people and businesses. Consumers are already using AI to help make financial decisions. A $2025$ survey found that $56\%$ of UK adults used AI to help manage their money in the previous year, where $37\%$ of those users used it for investment research and recommendations \citep{lloyds2025cdi}. In addition to user advice, Anthropic's Project Vend examined LLM management of a small shop \citep{anthropic2026deal,anthropic2025vend}. Salesforce's CRMArena-Pro analyzes agents performing business tasks \citep{huang2025crmarena}, Walmart uses automated supplier negotiation \citep{vanhoek2022walmart}, and Microsoft's Magentic Marketplace simulates customer and business agents negotiating and transacting \cite{bansal2025magentic}. As these interactions grow in frequency, deciding what information agents share becomes increasingly important. \looseness=-2

Economics has long studied the distinction between the private benefit of information and its social value. An optimal decision-maker cannot lose from more information because the strategy available with less information remains an option \citep{blackwell1953}. In strategic interactions, however, each agent's best choice depends on others' behavior, as captured by Nash equilibrium \citep{nash1950equilibrium}. Disclosure can therefore change equilibrium behavior, so information that benefits one agent need not improve collective outcomes \citep{hirshleifer1971,bergemann2016}. However, private information can also hinder the realization of mutually beneficial trades \citep{myerson1983efficient}. Information design studies how disclosure shapes these outcomes in Bayesian models \citep{kamenica2011,bergemann2019}. For LLM agents, how behavior and outcomes change under different information policies is an empirical question. \looseness=-2

Efforts to improve multi-agent LLM systems increasingly focus on optimizing their prompts and organization, as in GPTSwarm, MASS, and ROMA \citep{zhuge2024gptswarm,zhou2026mass,alzu2026roma}. We study a complementary choice of what information agents receive about others’ preferences, values, and costs, studying how disclosure shapes their decisions. Building on controlled negotiation environments such as NegotiationArena \citep{bianchi2024negotiationarena}, we compare disclosure policies on the same markets while holding models, incentives, task instructions, and interaction rules fixed.\footnote{We provide a detailed discussion of the related work in Appendix~\ref{app:extended-related}.} \looseness=-2

The results show that additional information does not always help the agents to reach better outcomes (Figure~\ref{fig:architecture}). In multi-resource trade, moving from coarse to full information disclosure lowers GPT-6 Sol's total gains from $97.7\%$ to $90.8\%$ of the maximum available, whereas GPT-6 Luna's rise from $68.9\%$ to $79.0\%$, showing different responses within the GPT family. Claude Sonnet 5 and Grok 4.3 also benefit from full disclosure, while Claude Opus 5.5 and Grok 4.7 achieve higher gains under coarse information. In bargaining, we assess whether agents reach agreement and how evenly they share the gains using the normalized Nash score, with disagreement scoring zero. At the six-message horizon shown in Figure~\ref{fig:architecture}, moving from coarse to full disclosure raises GPT-6 Sol's score from $0.71$ to $0.83$, but lowers GPT-5.1's from $0.77$ to $0.57$ and Grok 4.7's from $0.71$ to $0.17$. In auctions, full information lowers realized gains from $94.6\%$ to $64.1\%$ of the maximum available for Claude Opus 5.5 and from $97.4\%$ to $62.1\%$ for Grok 4.7. These comparisons use separate instances of the same model as interacting agents. As we show later, more information can also worsen outcomes in games between different model types. \looseness=-2

These results motivate \emph{decision-relevant sufficiency}, where agents need enough information to identify useful actions, with further detail judged by its effect on system outcomes. Building on this perspective, we make three main contributions:  
\vspace{-5pt}
\begin{itemize}[leftmargin=0.9cm]
    \item We present this framework as an evaluation environment to study the effect of information disclosure across three economic settings, holding models, markets, and interaction rules fixed.
    
     \vspace{-2pt}
         \item We analyze, across ten frontier language models, how disclosure affects performance and how its effects vary with the model, market structure, and interaction protocol.
    
     \vspace{-2pt}
     \item We distinguish the quality of trades agents identify from the outcomes they realize, examining their demands and the division of gains to interpret the gap.
\end{itemize}

\section{Experimental Setup}
\label{sec:arenas}

We begin by describing the three settings that are canonical in economics and capture features of real trade and negotiation, while allowing complex strategic behavior to emerge from simple rules. In all our experiments, agents are instructed to maximize their own gains, which reflects instances where they negotiate or trade on behalf of people or groups with unique interests. Throughout, we test ten models: GPT-5 mini, GPT-6 Luna, GPT-5.1, GPT-6 Sol, Gemini 3.7 Flash, Gemini 3.1 Pro, Claude Sonnet 5, Claude Opus 5.5, Grok 4.3, and Grok 4.7. For each, we use separate instances of that model as the interacting agents, each with its own role. In all settings, the information specified by each disclosure condition is provided in agents' prompts before they act. During trade and bargaining, agents may also choose what to reveal in conversation. Supplementary details can be found in \S\ref{app:protocols}. \looseness=-2

\begin{figure}[!ht]
  \centering
  \makebox[\textwidth][c]{%
  \includegraphics[width=\textwidth]{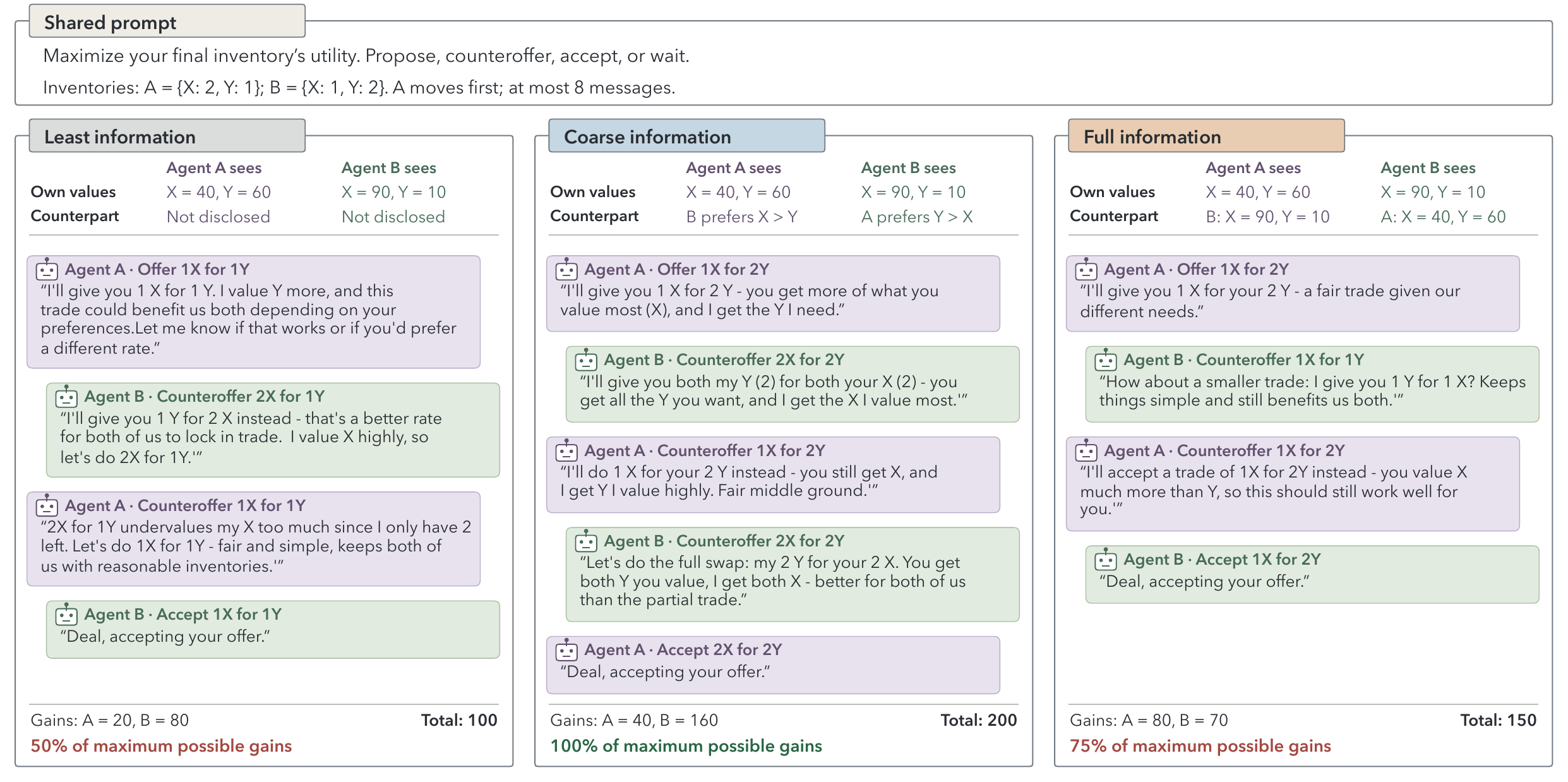}%
}
  \caption{\textbf{Disclosure changes negotiation in the same multi-resource exchange market.} Complete conversations between two Claude Sonnet 5 agents under least, coarse, and full information. The shared prompt is abbreviated; bold lines summarize actions. In this selected example, coarse information achieves the maximum possible total gain, compared with $50\%$ under least and $75\%$ under full information.}
  \label{fig:trade-conversation-example}
\end{figure}

\subsection{Multi-resource trade}

In this game, adapted from NegotiationArena \citep{bianchi2024negotiationarena}, two agents $A$ and $B$ negotiate a resource exchange with public inventories and private per-unit values. Both agents observe the inventories, and each knows its own values per resource unit. Utility is additive. If $t_i \in \mathbb{Z}^n$ lists the quantities of goods agent $i$ gives and $v_i \in \mathbb{R}^n$ their per-unit values, each agent's respective gain is:
\begin{equation}
g_A=v_A^\top(t_B-t_A),\qquad g_B=v_B^\top(t_A-t_B).
\label{eq:trade-gains}
\end{equation}
Agents alternate messages, starting with $A$, and may discuss, propose, counteroffer, accept, or wait. Offers can exchange any available integer quantities. Acceptance executes the exchange; without agreement, agents keep their initial inventories. The main study uses two resources, up to eight messages in total (four per agent), and $90$ markets\footnote{Markets specify the overall available resources and agents' values or costs. Averaging across markets helps us reduce variance and sensitivity to individual cases.} per model and information condition. Agents have complementary inventories and prefer the resource abundant in the counter-party's inventory. We vary preference strengths to create \emph{extraction opportunities}, where an agent can increase its own gain while reducing total gains and still benefiting its counter-party. Three groups, each consisting of $30$ markets give greater opportunity to A, to B, or equal opportunities, determined before model runs. \emph{See} \S\ref{app:trade-methods}. \looseness=-2

\noindent\textbf{Information levels.} We explore three information disclosure types. \Least{} provides only own resource values; \Coarse{} adds the counter-party's value ranking; and \Full{} adds its exact values. Inventories are public throughout, and agents choose what else to reveal in conversation. Figure~\ref{fig:trade-conversation-example} shows how the same market can produce different exchanges under different disclosure conditions. Under least information, the agents exchange one unit of each resource; under coarse information, they exchange two of each, doubling total gains. Under full information, A gives one X for two Y, increasing its own gain but reducing total gains relative to coarse disclosure.
\looseness=-2

\subsection{Single-good bargaining}

Bargaining provides a controlled setting for studying how information affects agreement and the division of gains \citep{rothmalouf1979}. In this game, a seller with private cost $c$ and a buyer with private value $v$ alternate price offers and responses. We test both starting roles, where the seller or the buyer agent starts first. An agreement at $c\le p\le v$ gives gains $p-c$ to the seller and $v-p$ to the buyer, assuming that $c, p$, and $v$ are all positive values. Total gains are therefore fixed at $v-c$; the price determines their division. We select $15$ profitable cost--value pairs, evenly divided among thin, medium, and large surplus markets, and test both starting roles at two, four, and six messages. The final message permits only acceptance or rejection, so two messages allow one offer and one response (\S\ref{app:bargaining-methods}). \looseness=-2

\noindent\textbf{Information levels.} Each agent knows its own value or cost and receives the prior that both are uniform on $[20,100]$. \Least{} provides no counter-party information; \Coarse{} reveals whether $v>c$, indicating whether a mutually beneficial price exists; and \Full{} reveals both cost and value. \looseness=-2

\subsection{Double auction}

Double auctions are a classical setting for studying how individual price demands produce market-level outcomes \citep{smith1962,gode1993}. Three buyers with private values $v_b$ and three sellers with private costs $c_s$ each seek to trade one unit of good.  In this one-shot auction, each buyer submits one bid ("what it offers to pay") and each seller one ask ("what it requests"), simultaneously and with no subsequent revision.
The \emph{clearing rule} ranks bids from highest to lowest and asks from lowest to highest, pairing participants in that order. A pair trades at the midpoint of its submitted prices if the bid meets or exceeds the ask. Clearing stops at the first pair that fails this condition. With $K$ completed trades between buyers $b(k)$ and sellers $s(k)$, total gain is:
\vspace{-4pt}
\begin{equation}
S_{\mathrm{auction}}=\sum\nolimits_{k=1}^{K}(v_{b(k)}-c_{s(k)}).
\label{eq:auction-welfare}
\end{equation}
We use $30$ markets varying in the number of profitable trades, whether the last profitable pair's gains are narrow, medium, or wide (\S\ref{app:auction-methods}). \looseness=-2

\noindent\textbf{Information levels.} Here, \Least{} provides each agent only its own value or cost, along with who the buyers and sellers are; \Coarse{} additionally reveals the ordering of the buyers by value and the sellers by cost; and \Full{} adds all exact values and costs. No generating distribution is disclosed.

\section{Measuring Discovery and Value Capture}
\label{sec:framework}\label{sec:common-result}

\begin{figure}[!t]
  \centering
  \includegraphics[width=\textwidth]{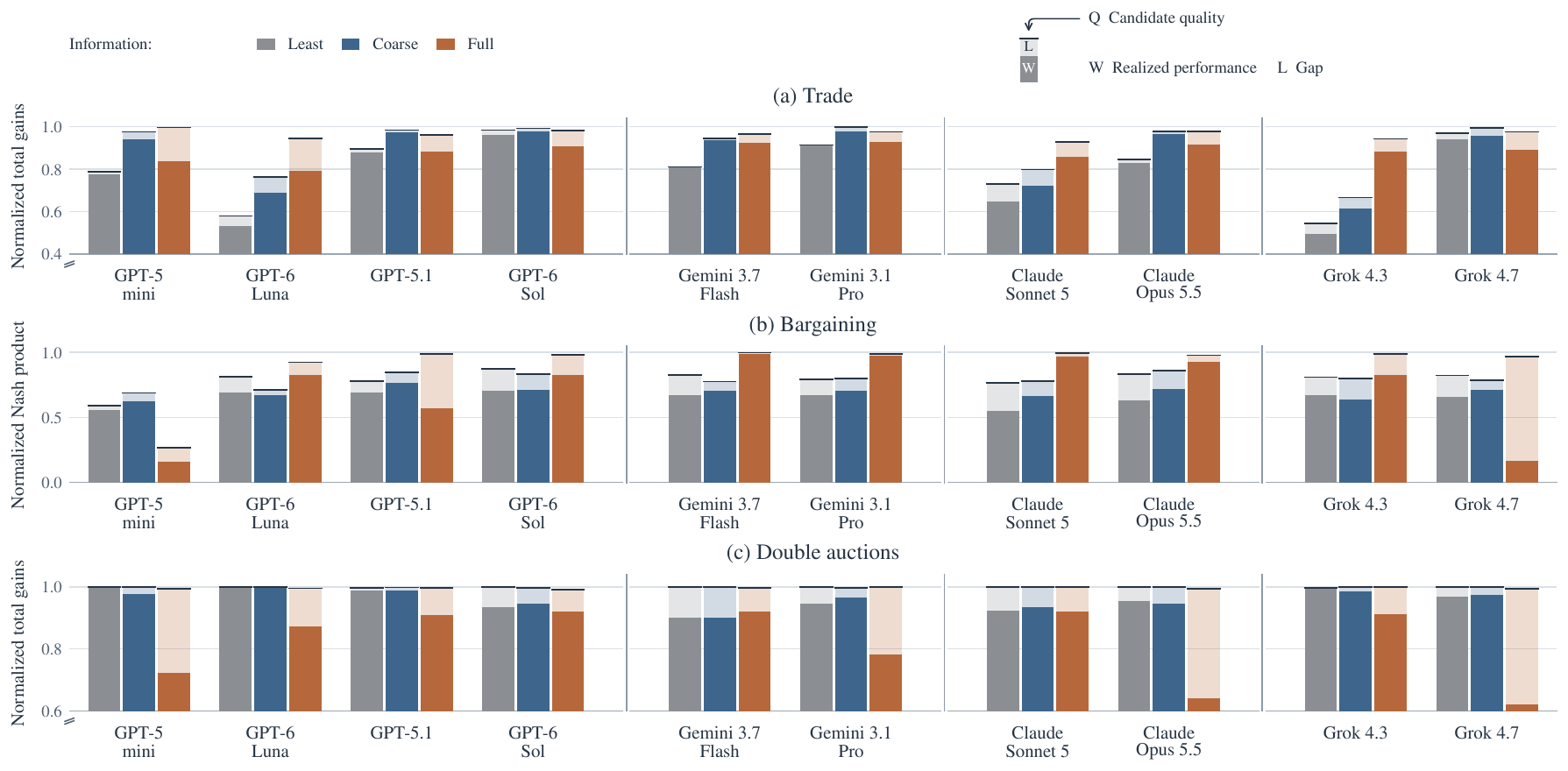}
\vspace{-8pt}
  \caption{\textbf{Candidate quality and realized performance across three economic settings.} Solid bars show realized performance $W$, the normalized score of completed outcomes. Total bar heights show candidate quality $Q$, the best feasible offer proposed in trade and bargaining, or the potential gains from auction pairs formed by submitted prices. The remaining segments show the realization gap $L=Q-W$, the difference between identified opportunities and realized outcomes. Scores measure total gains in trade and auctions and the Nash product in bargaining, with disagreements scored zero.}
  \label{fig:cross-arena-accounting}
\end{figure}

To understand how disclosure changes outcomes, we examine both the transactions identified by the agents and the agreements reached. Information may help agents find a more valuable trade whilst changing the terms they seek or accept. We compare the quality of the opportunities they identify with their final outcomes using the following three measures, as illustrated in Figure~\ref{fig:cross-arena-accounting}.

First, \textbf{\emph{realized performance}:} Let $S(a)$ score the system objective, $a^0$ denote no trade, and $a^*$ maximize the objective over feasible outcomes in the same market. We normalize $W(a)=[S(a)-S(a^0)]/[S(a^*)-S(a^0)]$ and call the final outcome’s score \emph{realized performance}, $W$. In trade and auctions, $S$ is total gains, so $W$ measures the proportion of maximum gains obtained. In bargaining, we use the normalized Nash product $W(p)=4(v-p)(p-c)/(v-c)^2$ to capture both agreement and balance, scoring equal division as one and disagreement as zero \citep{nash1950}. We average trade and bargaining negotiations equally, including disagreements, and weight auction markets by maximum available gains. We also report bargaining agreement rates and Nash scores among agreements. \looseness=-2

Second, \textbf{\emph{candidate quality}:} A good offer need not become the final agreement, reflecting distinctions between candidate generation and verification in LLMs \citep{saadfalcon2025shrinking}. \emph{Candidate quality}, $Q$, uses the same normalized objective as $W$. In trade and bargaining, it is the highest score among feasible proposed trades that leave both agents at least as well off as no trade, evaluated using true utilities. If no proposed trade satisfies these, we set $Q=0$. In auctions, it measures the normalized gains if every profitable pair formed by the submitted price rankings traded, even when the buyer offers less than what the seller asks. \looseness=-2

Third, \textbf{\emph{realization gap:}} The value $L=Q-W$ captures candidate quality \emph{not} realized in the final outcome. This can arise from accepting a lower-scoring offer or failing to complete beneficial trades. Using the same markets and weights, full-minus-coarse changes satisfy $\Delta W=\Delta Q-\Delta L$. We examine agents' demands, responses, and individual gains to interpret these differences.

\section{Main Results}
\label{sec:results}

We begin by comparing how disclosure affects the opportunities agents identify and the results they achieve. Figure~\ref{fig:cross-arena-accounting} summarizes candidate quality $Q$, realized performance $W$, and the realization gap $L$ across three settings. Overall, we see that the effects of information disclosure in economic exchanges differ across models and that full information often widens this realization gap. In trade, full disclosure lowers GPT-6 Sol's realized gains but raises GPT-6 Luna's, while Claude Sonnet 5 and Grok 4.3 improve both candidate quality and realized performance. In bargaining, Grok 4.7 and GPT-5.1 propose more balanced offers and have higher $Q$ under full information, but reach substantially less balanced outcomes. In auctions, the submitted prices can potentially pair buyers and sellers who would benefit from trading, yet prevent those trades because buyers offer less than sellers demand. To interpret these differences, we examine demands, responses, and individual gains; then consider how disclosure effects vary with market complexity, communication, and model composition. \looseness=-2

\begin{figure}[!t]
  \centering
  \includegraphics[width=\textwidth]{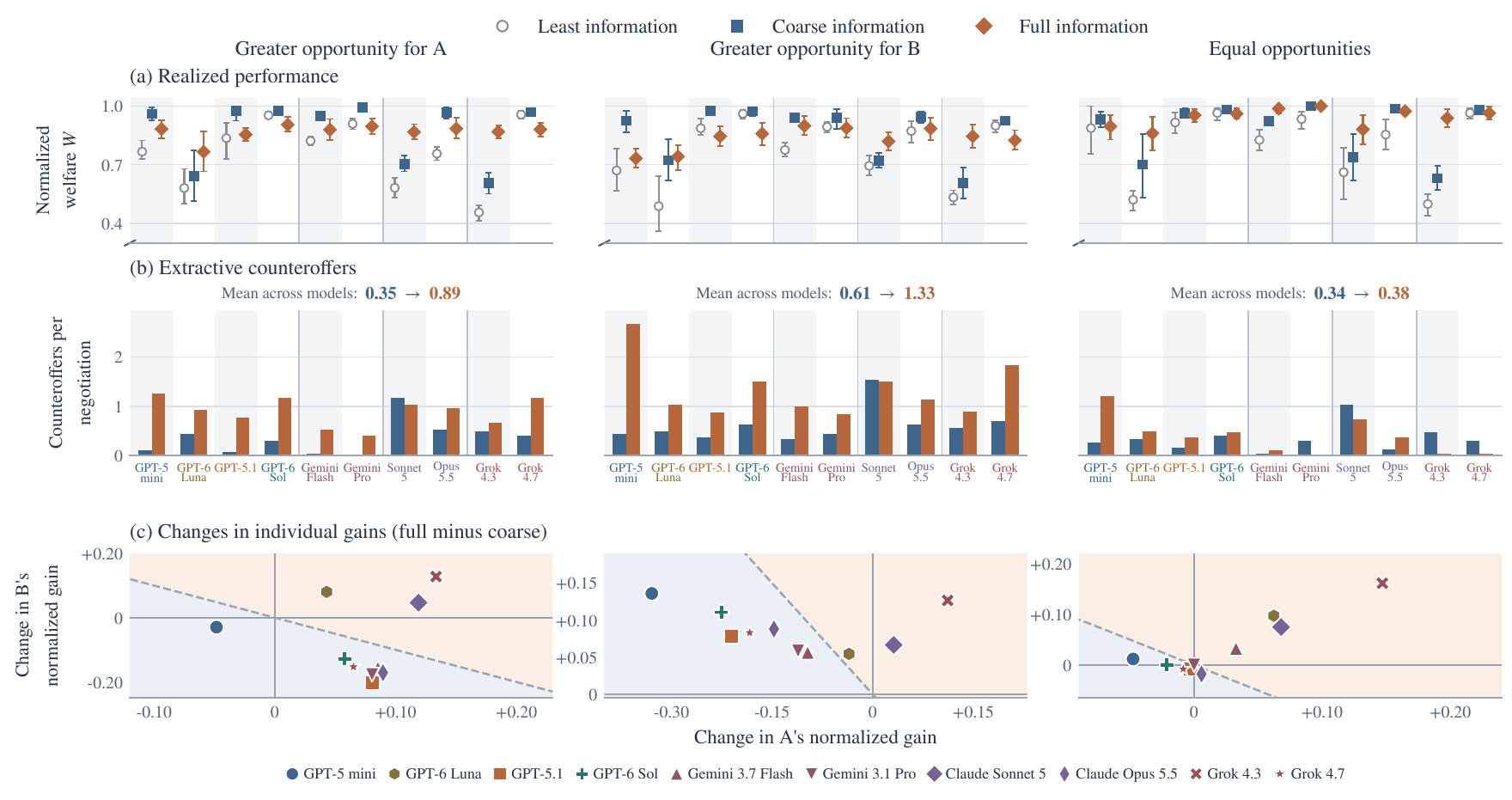}
  \vspace{-10pt}
  \caption{\textbf{Trade outcomes and behavior across market types.} Columns distinguish markets with greater extraction opportunity for A, for B, or equal opportunities. (a) Mean normalized welfare, with $95\%$ confidence intervals. (b) Showing mean \emph{extractive counteroffers} per conversation, which are counteroffers that increase the proposing agent's gain while reducing total gains relative to the preceding offer. (c) Each point shows a model's full-minus-coarse change in A's gain (horizontal) and B's gain (vertical), normalized by maximum total gain. Points above the dashed diagonal indicate higher total welfare. \emph{Together, the panels show that full disclosure can accompany more extractive counteroffers, less equal gains, and lower total welfare.} \looseness=-2}
  \label{fig:trade-outcomes-behavior}
  \label{fig:motivation}
  \label{fig:trade-capture}
\end{figure}

\subsection{Multi-resource trade}
\label{sec:trade-results}

Multi-resource trade requires agents to find an exchange and negotiate how its gains are divided. Disclosure can therefore affect both the trades they propose and the gains they ultimately achieve.

\begin{finding}{{Full information can sharpen capture while reducing total gains.}} In two-resource exchange, moving from counter-party value rankings to full information (exact values) lowers realized performance for GPT-6 Sol, Claude Opus 5.5, and Grok 4.7, and is accompanied by more extractive counteroffers.
\label{finding:trade}
\end{finding}

The aggregate effect differs across model families and market structures. Moving from coarse to full disclosure lowers GPT-6 Sol's realized gains from $97.7\%$ to $90.8\%$ of the maximum available, but raises GPT-6 Luna's from $68.9\%$ to $79.0\%$. GPT-5.1's gains fall from $97.2\%$ to $88.4\%$, whereas Claude Sonnet 5's rise from $72.1\%$ to $85.6\%$ (Figure~\ref{fig:cross-arena-accounting}). These consequences depend on how an agent can \emph{extract}, that is, improve its own gain while diminishing total profits but still with a positive gain offered for the other agent. In Figure~\ref{fig:motivation}(a), a total of $90$ markets are divided into three categories according to their opportunity of extraction: A, B or equal. Coarse disclosure generally produces higher total gains, including for GPT-6 Sol, Claude Opus 5.5 and Grok 4.7 in all three market groups. Full disclosure nevertheless benefits GPT-6 Luna, Claude Sonnet 5, and Grok 4.3 throughout. Gemini 3.7 Flash benefits from full information only when extraction opportunities are equal. \looseness=-2

We call a counteroffer \emph{extractive} when it increases the proposing agent's gain while reducing both the counter-party's gain and total gains relative to the previous offer. Figure~\ref{fig:trade-capture}(b) reports the average of such counteroffers per negotiation. We see that full disclosure increases extractive counteroffers for several models. When B has greater extraction opportunity, GPT-6 Sol's mean rises from $0.63$ to $1.50$ per negotiation, and Claude Opus 5.5's from $0.63$ to $1.13$. Such counteroffers can reduce final gains when accepted in place of an earlier, higher-value offer. However, their frequency alone does not determine performance. GPT-6 Luna makes more extractive counteroffers in every market group yet achieves higher overall gains, while Claude Sonnet 5 also benefits from improved proposals despite making extractive counteroffers under both conditions (Figure~\ref{fig:cross-arena-accounting}).

The changes in individual gains reveal who benefits from full disclosure. Figure~\ref{fig:trade-capture}(c) plots full-minus-coarse changes in A’s gain horizontally and B’s vertically, normalized by maximum total gains. Points below the diagonal indicate lower combined welfare. When B has greater extraction opportunity, GPT-6 Sol, Claude Opus 5.5, and Grok 4.7 increase B's gain while reducing A's by more. For GPT-6 Sol, B gains $11.1$ percentage points while A loses $22.5$. When A has greater opportunity, these models instead increase A's gain at a larger cost to B. By contrast, Claude Sonnet 5 and Grok 4.3 improve both agents' gains in all three market groups. With equal opportunities, Gemini 3.7 Flash also improves both agents' gains despite making more extractive counteroffers. Full information can therefore benefit one agent at a greater cost to the other, reducing their combined gains. \looseness=-2

\subsection{Single-good bargaining}
\label{sec:bargaining-results}

Within each bargaining market, every completed trade creates the same total gains. We therefore examine whether agents agree and how evenly they divide those gains. Here, $Q$ scores the most balanced feasible price proposed, while $W$ scores the final agreement and is zero for disagreements.

\begin{finding}{Disclosure can increase agreement \emph{while} {worsening} the division of gains.} Full information disclosure can make agreement more plausible, yet fail to produce a correspondingly balanced allocation of the surplus available. 
\label{finding:bargaining}
\end{finding}

\begin{figure}[tbp]
  \centering
  \includegraphics[width=\textwidth]{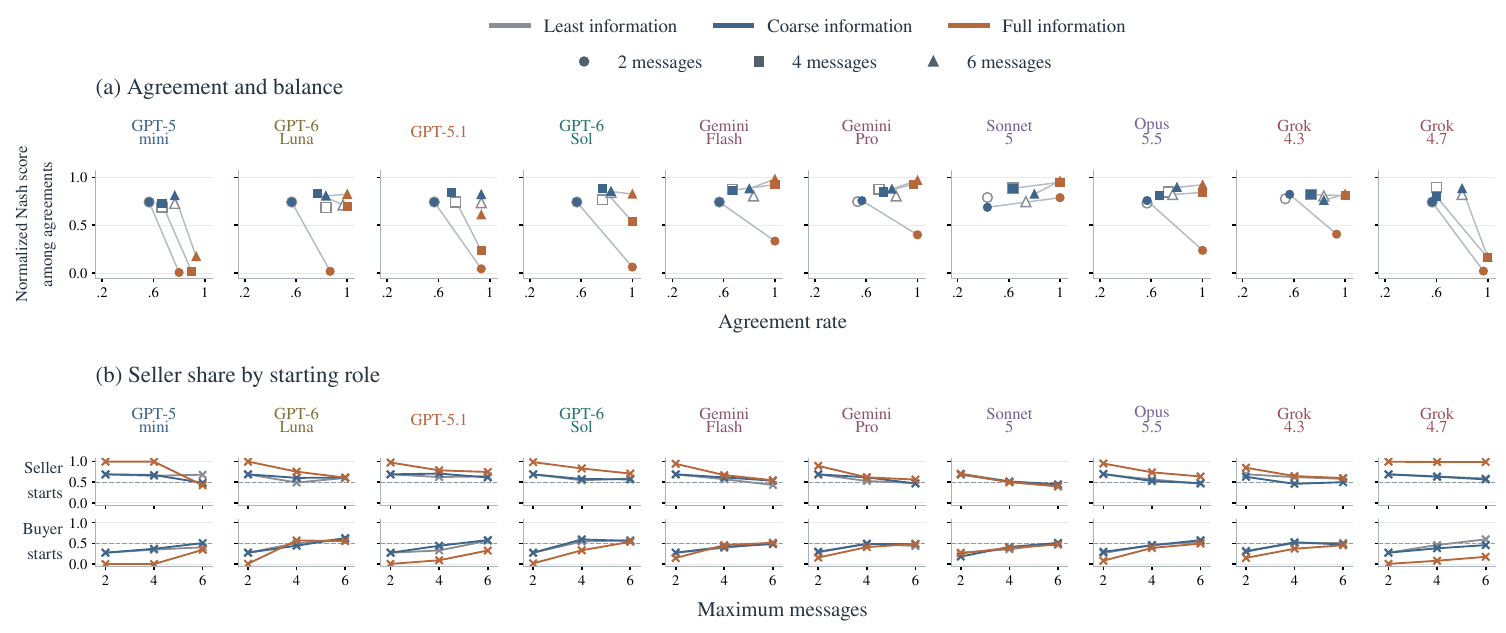}
  \vspace{-18pt}
  \caption{\textbf{Bargaining agreement, balance, and starting-role effects.} (a) Agreement rates versus mean normalized Nash products among agreements, combining experiments with both starting roles. Shapes indicate message limits; gray lines connect coarse and full information at the same limit. (b) Mean seller shares among agreements separated by starting role, and dashed lines mark equal division. \emph{More information can increase agreement while making gains less evenly shared, with the division also depending on who moves first.} \looseness=-2}%
  \label{fig:bargaining-combined}
  \label{fig:bargaining-horizon-app}
  \label{fig:bargaining-surplus}
\end{figure}

Figure~\ref{fig:bargaining-combined}(a) illustrates the difference between agreement and balance, where points farther to the right suggest more agreement, but points higher up imply more balanced agreement. With four messages, full disclosure raises GPT-5.1’s agreement rate from $70\%$ to $93.3\%$ but lowers the mean Nash score among agreements from $0.84$ to $0.24$. Grok 4.7 shows a similar pattern at six messages, where the agreement rises from $80\%$ to $100\%$, while the Nash score among agreements falls from $0.89$ to $0.17$. Its candidate quality nevertheless improves from $0.79$ to $0.97$, showing that balanced offers appear without becoming balanced final agreements. \looseness=-2

Longer dialogue improves balance under full information for several models. Between two and six messages, scores among agreements rise from $0.02$ to $0.83$ for GPT-6 Luna, from $0.06$ to $0.83$ for GPT-6 Sol, and from $0.24$ to $0.92$ for Claude Opus 5.5. Gemini 3.7 Flash, Claude Sonnet 5, and Grok 4.3 also reach more balanced agreements with longer dialogue. At six messages, full disclosure improves overall realized performance for seven models, but lowers it for GPT-5 mini, GPT-5.1, and Grok 4.7. Results by surplus group appear in \S\ref{app:bargaining-results}. \looseness=-2

\begin{finding}{Information and bargaining order jointly shape who captures the surplus.} Under short bargaining horizons, full information amplifies the first-mover's share of gains.
\label{finding:information-starting-role}
\end{finding}
Figure~\ref{fig:bargaining-surplus}(b) shows the mean shares of the sellers among the agreements, separated by the starting roles. With two messages, moving from coarse to full disclosure raises the GPT-5.1 sellers' share from $68.8\%$ to $97.3\%$ when starting, but reduces it from $27.4\%$ to $0.4\%$ when responding. At six messages under full information, these shares are $74.5\%$ and $32.6\%$, narrowing the starting role advantage. Grok 4.3, Gemini 3.7 Flash, and Claude Sonnet 5 also reach more balanced agreements as dialogue length increases under full information. By contrast, Grok 4.7 retains a strong starting role advantage even at six messages under full information, such that sellers receive $98.7\%$ when starting and $17.4\%$ when responding. Starting order also matters when only one of the agents receives exact information, as an informed second mover can still receive less than its uninformed counterpart (\S\ref{app:bargaining-results}). Together, these findings show that information and starting role jointly shape who gains. Human bargaining experiments likewise show that disclosure changes the agreed division \citep{rothmalouf1979,rothmalouf1982} and that its recipient matters \citep{rothmurnighan1982}, although the direction of disclosure effects varies across our tested models. \looseness=-2

\subsection{Double auctions}
\label{sec:auction-results}
In auctions, disclosure affects both who is matched with whom and whether their prices allow trade.

\begin{finding}{Agents can identify efficient trades but still price them out of existence.}
Full information can prevent mutually beneficial trades in auctions, making the prices buyers offer and sellers demand less compatible. In other words, even though the exchange may benefit both parties, agents may complete fewer trades and achieve lower total gains.
\label{finding:auction}
\end{finding}

\begin{figure}[tbp]
  \centering
\includegraphics[width=\textwidth]{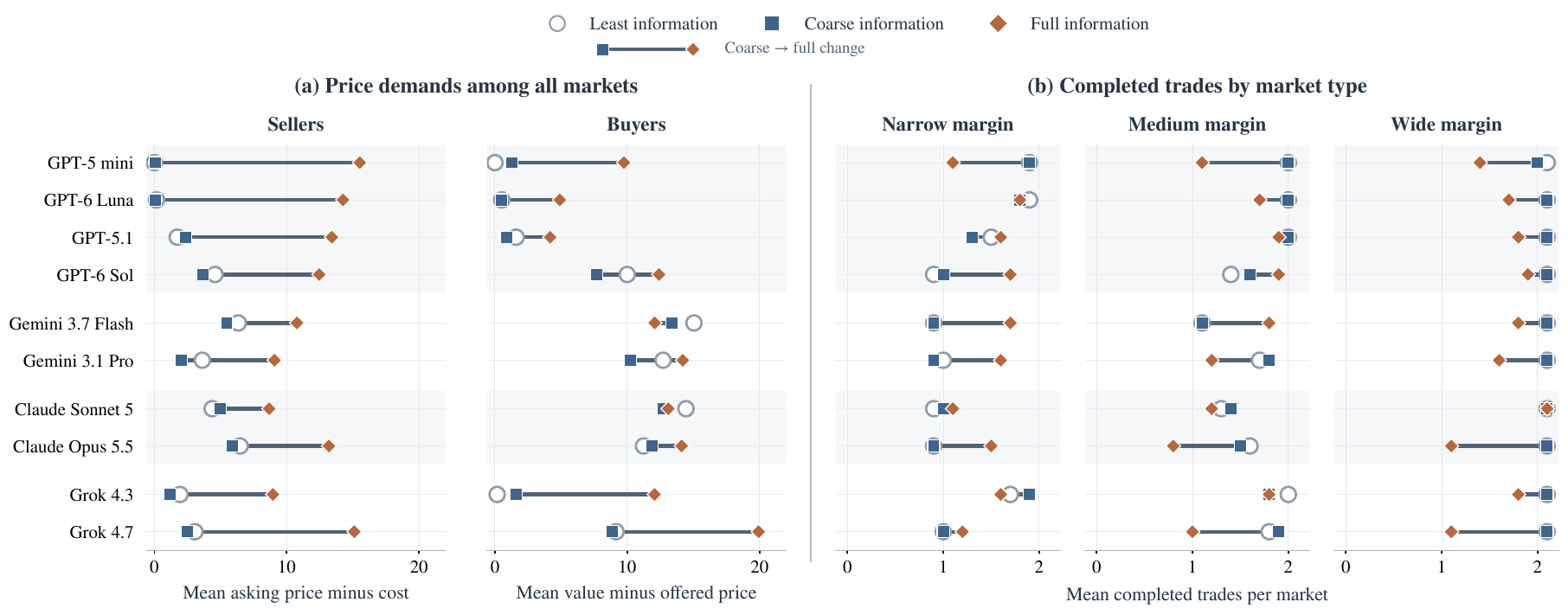}
\vspace{-10pt}
    \caption{\textbf{Auction price demands and completed trades.} \textit{Full information can lead sellers to ask for more and buyers to offer less, preventing trades that would benefit both especially in wider margin markets.} (a) Mean asking prices above seller costs and mean buyer values minus offered prices across all market types.
    (b) Mean completed trades in narrow, medium, and wide margin markets. Symbols distinguish information conditions and lines connect coarse and full information.}
  \label{fig:auction-behavior}
\end{figure}

The auction clearing rule produces buyer--seller pairs that could generate nearly all available gains in our experiments, resulting in high candidate quality $Q$ (Figure~\ref{fig:cross-arena-accounting}). But these gains disappear when buyers are willing to pay less than what the sellers ask. For GPT-5.1, the pairs could generate over $99.7\%$ of maximum gains under either condition, but full disclosure lowers realized gains from $98.9\%$ to $91.0\%$. The declines are larger for Claude Opus 5.5 and Grok 4.7, whose realized gains fall from $94.6\%$ to $64.1\%$ and from $97.4\%$ to $62.1\%$, respectively, despite candidate quality remaining above $99\%$. These models complete fewer beneficial trades, widening the realization gap $L$. Claude Sonnet shows a smaller decline in realized gains, from $93.6\%$ to $92.0\%$. 

Figure~\ref{fig:auction-behavior}(a) helps explain why some gains remain unrealized. Compared with coarse information, full disclosure leads sellers to ask for more above their costs in every model. Buyers also offer further below their values in every model except Gemini 3.7 Flash. For clarity, price differences are capped at $100$ before averaging to remove extreme values. Furthermore, auction activity depends strongly on the market structure (Figure~\ref{fig:auction-behavior} b). Full disclosure increases mean completed trades in narrow-margin markets for several models, including GPT-6 Sol and Gemini 3.7 Flash; but reduces them in wide-margin markets for nine models, with Claude Sonnet 5 unchanged. In these wide-margin markets, even the last efficient trade offers substantial gains, making missed trades particularly consequential.

\subsection{Additional Analysis and Ablations} 
\label{sec:ablations}

\textit{\textbf{Resource count and communication.}} We examine how the value of information varies with the number of resources and conversation length, using markets where B has greater extraction opportunity. \looseness=-2 %

\begin{finding}{Trading complexity changes the value of information disclosure.}
As agents face more possible exchanges, exact preferences improve the trade discovery enough to shift the advantage from coarse to full information.
\label{finding:resource-discovery}
\end{finding}

In the two-resource experiments, coarse information appears sufficient to identify high-value trades, leaving less room to improve proposals. When agents face combinatorially more possible exchanges, however, detailed preferences can help to find better trades. At ten resources, full information improves over coarse information candidate quality $Q$ for all ten models and realized performance $W$ for nine, with GPT-6 Sol nearly unchanged. GPT-5.1's realized gains rise from $63.8\%$ to $80.4\%$ of the maximum available. Benefits vary with resource count, which also changes the available exchanges (Figure~\ref{fig:trade-resource-boundary}(a)). \looseness=-2

Beyond resource count, communication also shapes the value of disclosure. Extending conversations from two to twelve messages improves welfare more under least than full information for most models. With more communication, GPT-5.1 gains $30.6$ percentage points under least information versus $2.8$ under full disclosure, suggesting that dialogue can partly compensate for limited initial information (Figure~\ref{fig:trade-horizons-app}(b)). We also examine a mediator setup that maximizes total gains, subject to both agents’ acceptance (\cref{fig:trade-mediator}). In that setup, full information yields near-optimal outcomes, while direct negotiation favors coarse disclosure on the same markets, showing that information’s value depends on objectives and the interaction architecture.

\begin{figure}[!t]
  \centering
  \includegraphics[width=\textwidth]{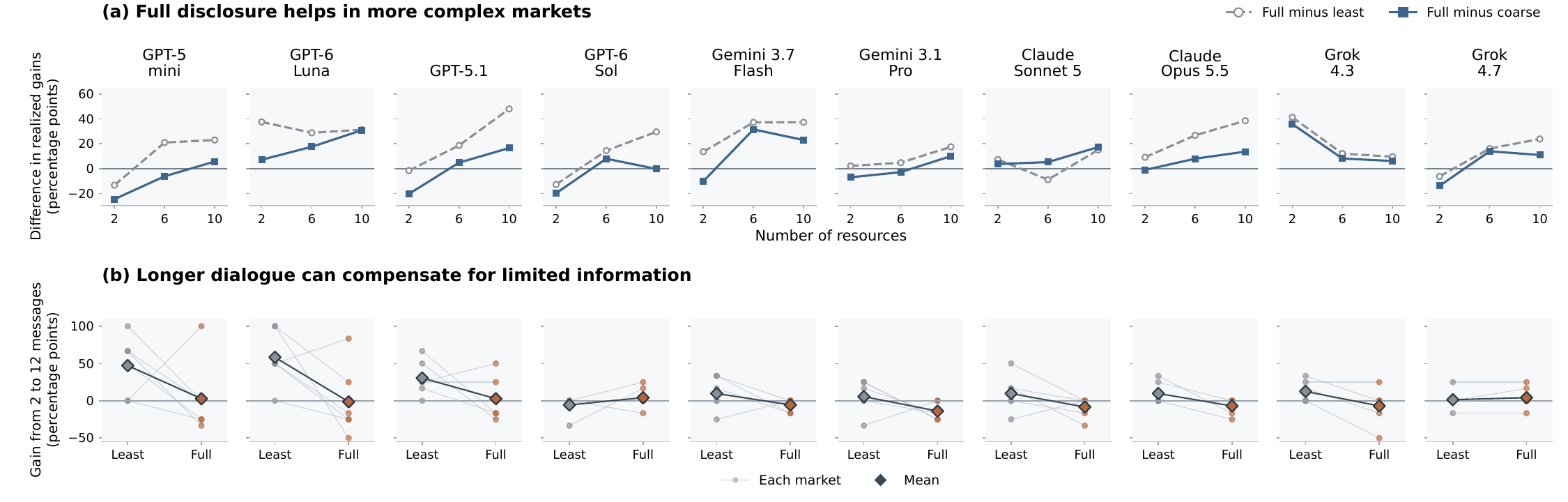}
  \vspace{-15pt}
  \caption{\textbf{Resource complexity and communication.} (a) Differences in realized gains between full and least or coarse information at each resource count. Positive values favor full disclosure. (b) Changes in realized gains from two to twelve messages under least and full information. Lines connect the same markets and diamonds show their means. All differences are in percentage points of maximum possible gains. Both studies use markets where B has greater extraction opportunity. \looseness=-2}
  \label{fig:trade-sweeps}
  \label{fig:trade-resource-boundary}
  \label{fig:trade-horizons-app}
\end{figure}

\textit{\textbf{Mixed-model interactions.}} To see if these trends also hold for interactions between different models, we test all pairs of GPT-5.1, Claude Sonnet 5, Gemini 3.7 Flash, and Grok 4.3 on identical markets used in previous experiments. The full disclosure is compared with coarse in Figure~\ref{fig:mixed-model-effects}, averaging both model-role assignments. The bargaining employs six messages and both starting roles. The trade results echo the same-model findings. GPT-5.1 $\times$ Gemini loses $9.6\%$ points in each unequal-opportunity group, with little change under equal opportunities, while Sonnet $\times$ Grok gains $10.8$--$23.8$ percentage points across groups. In bargaining, the pairs including GPT-5.1 retain the disadvantage of full disclosure in large-surplus markets, losing $13.3$--$28.4$ points, although every pair benefits when surplus is thin. In auctions, all six model combinations achieve higher mean performance under coarse information, with full disclosure reducing scores by $4.9$--$14.6\%$ points. Coarse information also performs better for every combination in wide-margin markets, consistent with the same-model findings. These results show that several disclosure effects persist when different models interact, while their direction and magnitude depend on the models involved and the market (\S\ref{app:mixed-models}). \looseness=-2

\label{sec:mixed-models}

\begin{figure}[tbp]
  \centering
  \includegraphics[width=\textwidth]{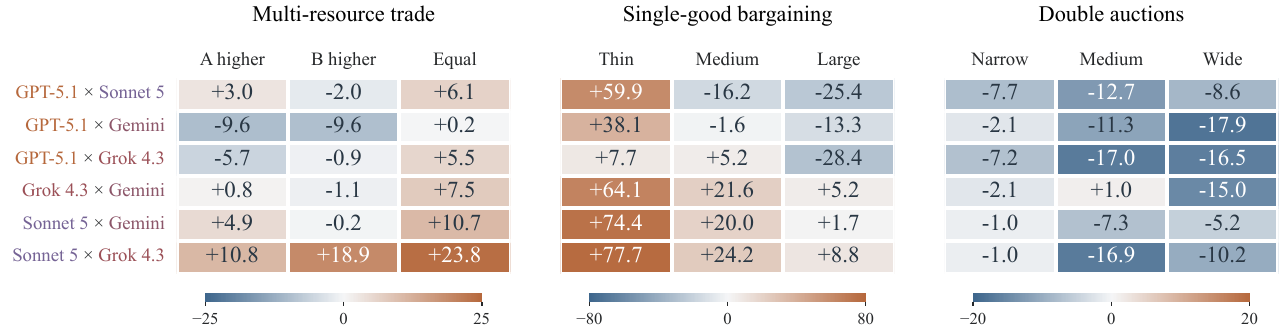}
  \vspace{-12pt}
  \caption{\textbf{Disclosure effects in mixed-model interactions.} Mean change in realized performance $W$ from coarse to full disclosure (percentage points). \textbf{Blue favors coarse disclosure; orange favors full}. Columns group markets by extraction opportunity, bargaining surplus, or auction margin. Trade and auctions measure normalized total gains and bargaining measures the normalized Nash product.}
  \label{fig:mixed-model-effects}
\end{figure}

\textit{\textbf{Strategic demands and inconsistent behavior.}} Conversations reveal interesting behavior in negotiations. In six-message bargaining, GPT-5.1 makes its last possible offer on the fifth message in $56.7\%$ of same-model negotiations under full information, compared with $16.7\%$ under coarse information. Among these offers, it demands a larger share than in its previous proposal in $76.5\%$ and $40\%$ of cases, for full and coarse information respectively. Across all message limits, Claude Sonnet 5 attempts to accept offers that would leave it worse off than not trading in $8.9\%$ of least-information and $5.6\%$ of coarse-information negotiations, but none under full information. Multi-resource trade also reveals misleading claims and successful error correction. GPT-5.1 calls a counteroffer "\textit{more balanced}" while raising its own share from $26.7\%$ to $93.3\%$. Then, its counter-party objects to this, leading to a more balanced agreement. In another negotiation, Gemini challenges GPT-5.1's claim that no beneficial trade exists and secures an exchange achieving the maximum total gain possible. These cases illustrate how correcting a counter-party can improve outcomes. Further details on these conversations appear in \S\ref{app:output-consistency}. \looseness=-2
\label{sec:output-consistency}

\section{Discussion, Limitations, and Conclusion}
\label{sec:synthesis}\label{sec:limitations}\label{sec:conclusion}

The results in LLM-agent markets depend on the models, interaction rules, and the information shared. Too little information can hinder beneficial trades, while additional detail can intensify competition over gains, making the appropriate level of disclosure important. Better proposals need not increase total gains, nor does more frequent agreement guarantee a more equal distribution of those gains. Comparing candidate quality, agents' demands, and realized performance helps explain these differences. Model selection and information policy should therefore be considered together, based on how agents use information and whether their interactions support the system’s objective. \looseness=-2

\textbf{Implications and future directions.} Information availability is an important consideration when developing objectives, decision authority, and interaction rules for agents representing humans or organizations. Commercial agents may have varying levels of access to client needs, supplier capabilities, and operational constraints depending on their role. Similar decisions should be made about context and feedback in coding and planning systems, but further validation is required for transfer outside of economic contexts. Our dialogue length and resource count studies encourage testing when to provide detail first and when to reveal it through interaction. Future research could look at selective disclosure in unseen markets and model pairs, measuring both collective outcomes and gain allocation. \looseness=-2

\textbf{Limitations.} Our experiments isolate disclosure effects in controlled games with known utilities and explicit objectives, whereas real-world agents may face uncertain preferences and changing goals. Findings may also vary across economic settings, models, and interaction rules. The markets we considered are selected to capture specific trading characteristics rather than the full range of real-world interactions. Furthermore, our confidence intervals capture variation across markets, not repeated model responses to the same input. Accordingly, studies using real-world data and deployed agents are needed to assess how these findings generalize. \looseness=-2

\section*{Acknowledgements}

We would like to thank the members of the Zou Lab for helpful discussions and valuable feedback. Suzgun gratefully acknowledges the support of a Google PhD Fellowship.
Pappu and El gratefully acknowledge the support of the Knight-Hennessy Scholarship. 
This work was further supported in part by the National Science Foundation (NSF) CAREER Award under Grant CCF-2236829, in part by the NSF Grant CPS-2551283, in part by the National Institutes of Health under Grant 1R01AG08950901A1, in part by the Office of Naval Research under Grant N00014-24-1-2164, and in part by the Defense Advanced Research Projects Agency under Grant HR00112490441.

\section*{AI Use Statement}
We used generative AI tools, specifically ChatGPT  and Claude, to assist with writing and editing, implementing code, processing and organizing data, and interpreting quantitative results. 

Market instances followed explicit rules rather than free-form LLM responses. The LLM agents evaluated in the experiments and their configurations are described separately in the paper. This work contains no mathematical proofs. We reviewed the AI-assisted work and take responsibility for the final text, code, and reported findings. \looseness=-2

\FloatBarrier
\setlength{\parskip}{0.5pc}
\begingroup
\small
\bibliography{references}

@InProceedings{bianchi2024negotiationarena,
  title = 	 {How Well Can {LLM}s Negotiate? {N}egotiation{A}rena Platform and Analysis},
  author =       {Bianchi, Federico and Chia, Patrick John and Yuksekgonul, Mert and Tagliabue, Jacopo and Jurafsky, Dan and Zou, James},
  booktitle = 	 {Proceedings of the 41st International Conference on Machine Learning},
  pages = 	 {3935--3951},
  year = 	 {2024},
  editor = 	 {Salakhutdinov, Ruslan and Kolter, Zico and Heller, Katherine and Weller, Adrian and Oliver, Nuria and Scarlett, Jonathan and Berkenkamp, Felix},
  volume = 	 {235},
  series = 	 {Proceedings of Machine Learning Research},
  month = 	 {21--27 Jul},
  publisher =    {PMLR},
  url = 	 {https://proceedings.mlr.press/v235/bianchi24a.html}
}

@article{bergemann2019,
  title={Information design: A unified perspective},
  author={Bergemann, Dirk and Morris, Stephen},
  journal={Journal of economic literature},
  volume={57},
  number={1},
  pages={44--95},
  year={2019},
  publisher={American Economic Association 2014 Broadway, Suite 305, Nashville, TN 37203-2425}
}

@article{glode2018,
  title={Voluntary disclosure in bilateral transactions},
  author={Glode, Vincent and Opp, Christian C and Zhang, Xingtan},
  journal={Journal of Economic Theory},
  volume={175},
  pages={652--688},
  year={2018},
  publisher={Elsevier}
}

@inproceedings{riedl2026,
  title={Emergent coordination in multi-agent language models},
  author={Riedl, Christoph},
  booktitle={International Conference on Learning Representations},
  volume={2026},
  pages={120776--120799},
  year={2026}
}

@misc{anthropic2026multiagent,
  title        = {Patterns and Problems in Emerging Multiagent Systems},
  author       = {{Anthropic}},
  year         = {2026},
  howpublished = {Research blog, August 13},
  url          = {https://www.anthropic.com/research/multiagent-systems}
}

@misc{anthropic2025vend,
  title        = {Project Vend: Can {Claude} Run a Small Shop? (And Why Does That Matter?)},
  author       = {{Anthropic}},
  year         = {2025},
  month        = jun,
  howpublished = {Research blog, June 27},
  url          = {https://www.anthropic.com/research/project-vend-1}
}

@misc{anthropic2026deal,
  author       = {Troy, Kevin K. and Shields, Dylan and Bradwell, Keir and McCrory, Peter},
  title        = {{Project Deal}: Our {Claude}-Run Marketplace Experiment},
  howpublished = {Anthropic},
  year         = {2026},
  month        = apr,
  day          = {24},
  url          = {https://www.anthropic.com/features/project-deal},
  note         = {Accessed September 23, 2026}
}

@article{arike2026,
  title={How does information access affect LLM monitors' ability to detect sabotage?},
  author={Arike, Rauno and Moreno, Raja Mehta and Subramani, Rohan and Biswas, Shubhorup and Ward, Francis Rhys},
  journal={arXiv preprint arXiv:2601.21112},
  year={2026}
}

@article{akinwande2026,
  title={Sharding Prevents LLM Oversight Failures and Adversarial Exploitation},
  author={Akinwande, Victor and Kolter, J Zico and Nayebi, Aran},
  journal={arXiv preprint arXiv:2608.06422},
  year={2026}
}

@article{blackwell1953,
  title={Equivalent comparisons of experiments},
  author={Blackwell, David},
  journal={The annals of mathematical statistics},
  pages={265--272},
  year={1953},
  publisher={JSTOR}
}

@incollection{hirshleifer1971,
  title={The private and social value of information and the reward to inventive activity},
  author={Hirshleifer, Jack},
  booktitle={Uncertainty in economics},
  pages={541--556},
  year={1978},
  publisher={Elsevier}
}

@article{kamenica2011,
  title={Bayesian persuasion},
  author={Kamenica, Emir and Gentzkow, Matthew},
  journal={American Economic Review},
  volume={101},
  number={6},
  pages={2590--2615},
  year={2011},
  publisher={American Economic Association}
}

@article{bergemann2016,
  title={Bayes correlated equilibrium and the comparison of information structures in games},
  author={Bergemann, Dirk and Morris, Stephen},
  journal={Theoretical Economics},
  volume={11},
  number={2},
  pages={487--522},
  year={2016},
  publisher={Wiley Online Library}
}

@article{morris2002,
  title={Social value of public information},
  author={Morris, Stephen and Shin, Hyun Song},
  journal={american economic review},
  volume={92},
  number={5},
  pages={1521--1534},
  year={2002},
  publisher={American Economic Association}
}

@article{crawford1982,
  title={Strategic information transmission},
  author={Crawford, Vincent P and Sobel, Joel},
  journal={Econometrica: Journal of the Econometric Society},
  pages={1431--1451},
  year={1982},
  publisher={JSTOR}
}

@article{chatterjee1983,
  title={Bargaining under incomplete information},
  author={Chatterjee, Kalyan and Samuelson, William},
  journal={Operations research},
  volume={31},
  number={5},
  pages={835--851},
  year={1983},
  publisher={INFORMS}
}

@article{smith1962,
  title={An experimental study of competitive market behavior},
  author={Smith, Vernon L},
  journal={Journal of political economy},
  volume={70},
  number={2},
  pages={111--137},
  year={1962},
  publisher={The University of Chicago Press}
}

@inproceedings{lewis2017,
  title={Deal or no deal? end-to-end learning of negotiation dialogues},
  author={Lewis, Mike and Yarats, Denis and Dauphin, Yann and Parikh, Devi and Batra, Dhruv},
  booktitle={Proceedings of the 2017 conference on empirical methods in natural language processing},
  pages={2443--2453},
  year={2017}
}

@inproceedings{he2018,
  title={Decoupling strategy and generation in negotiation dialogues},
  author={He, He and Chen, Derek and Balakrishnan, Anusha and Liang, Percy},
  booktitle={Proceedings of the 2018 Conference on Empirical Methods in Natural Language Processing},
  pages={2333--2343},
  year={2018}
}

@article{akata2023,
  title={Playing repeated games with large language models},
  author={Akata, Elif and Schulz, Lion and Coda-Forno, Julian and Oh, Seong Joon and Bethge, Matthias and Schulz, Eric},
  journal={Nature Human Behaviour},
  volume={9},
  number={7},
  pages={1380--1390},
  year={2025},
  publisher={Nature Publishing Group UK London}
}

@article{li2023camel,
  title={Camel: Communicative agents for" mind" exploration of large language model society},
  author={Li, Guohao and Hammoud, Hasan and Itani, Hani and Khizbullin, Dmitrii and Ghanem, Bernard},
  journal={Advances in neural information processing systems},
  volume={36},
  pages={51991--52008},
  year={2023}
}

@article{wu2023autogen,
  title   = {{AutoGen}: Enabling Next-Gen {LLM} Applications via Multi-Agent Conversation},
  author  = {Wu, Qingyun and Bansal, Gagan and Zhang, Jieyu and Wu, Yiran and Li, Beibin and Zhu, Erkang and Jiang, Li and Zhang, Xiaoyun and Zhang, Shaokun and Liu, Jiale and Awadallah, Ahmed Hassan and White, Ryen W. and Burger, Doug and Wang, Chi},
  journal = {arXiv preprint arXiv:2308.08155},
  year    = {2023}
}

@article{zhang2026termsbench,
  title={TERMS-Bench: Diagnosing LLM Negotiation Agents Beyond Deal Rate},
  author={Zhang, Erica and Zhang, Fangzhao and Pappu, Aneesh and El, Batu and Blanchet, Jose and Athey, Susan and Liu, Jiashuo and Zou, James},
  journal={arXiv preprint arXiv:2605.13909},
  year={2026}
}

@article{yashwanth2026sotopiatom,
  title={SOTOPIA-TOM: Evaluating Information Management in Multi-Agent Interaction with Theory of Mind.},
  author={Yashwanth, YS and Wang, Ruichen and Zeng, Shihua and Zhou, Xuhui and Onoue, Koichi and Varadarajan, Vasudha and Sap, Maarten},
  journal={Preprint},
  year={2026}
}

@article{rubinstein1982,
  title={Perfect equilibrium in a bargaining model},
  author={Rubinstein, Ariel},
  journal={Econometrica: Journal of the Econometric Society},
  pages={97--109},
  year={1982},
  publisher={JSTOR}
}

@article{nash1950,
  title={The bargaining problem},
  author={Nash, John F},
  journal={Econometrica},
  volume={18},
  number={2},
  pages={155--162},
  year={1950}
}

@article{gode1993,
  title={Allocative efficiency of markets with zero-intelligence traders: Market as a partial substitute for individual rationality},
  author={Gode, Dhananjay K and Sunder, Shyam},
  journal={Journal of political economy},
  volume={101},
  number={1},
  pages={119--137},
  year={1993},
  publisher={The University of Chicago Press}
}

@article{rothmalouf1979,
  title={Game-theoretic models and the role of information in bargaining.},
  author={Roth, Alvin E and Malouf, Michael W},
  journal={Psychological review},
  volume={86},
  number={6},
  pages={574},
  year={1979},
  publisher={American Psychological Association}
}

@article{rothmalouf1982,
  title={Scale changes and shared information in bargaining: An experimental study},
  author={Roth, Alvin E and Malouf, Michael WK},
  journal={Mathematical Social Sciences},
  volume={3},
  number={2},
  pages={157--177},
  year={1982},
  publisher={Elsevier}
}

@article{rothmurnighan1982,
  title={The role of information in bargaining: An experimental study},
  author={Roth, Alvin E and Murnighan, J Keith},
  journal={Econometrica: Journal of the Econometric Society},
  pages={1123--1142},
  year={1982},
  publisher={JSTOR}
}

@inproceedings{deng2024bargaining,
  title={Llms at the bargaining table},
  author={Deng, Yuan and Mirrokni, Vahab and Leme, Renato Paes and Zhang, Hanrui and Zuo, Song},
  booktitle={Agentic Markets Workshop at ICML},
  volume={2024},
  year={2024}
}

@inproceedings{xia2024bargaining,
  title={Measuring bargaining abilities of llms: A benchmark and a buyer-enhancement method},
  author={Xia, Tian and He, Zhiwei and Ren, Tong and Miao, Yibo and Zhang, Zhuosheng and Yang, Yang and Wang, Rui},
  booktitle={Findings of the Association for Computational Linguistics: ACL 2024},
  pages={3579--3602},
  year={2024}
}

@article{shapira2024glee,
  title={Glee: A unified framework and benchmark for language-based economic environments},
  author={Shapira, Eilam and Madmon, Omer and Reinman, Itamar and Amouyal, Samuel Joseph and Reichart, Roi and Tennenholtz, Moshe},
  journal={arXiv preprint arXiv:2410.05254},
  year={2024}
}

@article{chen2023aucarena,
  title={Put your money where your mouth is: Evaluating strategic planning and execution of llm agents in an auction arena},
  author={Chen, Jiangjie and Yuan, Siyu and Ye, Rong and Majumder, Bodhisattwa Prasad and Richardson, Kyle},
  journal={arXiv preprint arXiv:2310.05746},
  year={2023}
}

@article{fish2024collusion,
  title={Algorithmic collusion by large language models},
  author={Fish, Sara and Gonczarowski, Yannai A and Shorrer, Ran I},
  journal={arXiv preprint arXiv:2404.00806},
  volume={7},
  number={2},
  pages={5},
  year={2024},
  publisher={Working paper}
}

@inproceedings{duetting2024mechanism,
  title={Mechanism design for large language models},
  author={Duetting, Paul and Mirrokni, Vahab and Paes Leme, Renato and Xu, Haifeng and Zuo, Song},
  booktitle={Proceedings of the ACM Web Conference 2024},
  pages={144--155},
  year={2024}
}

@inproceedings{dubey2024summaries,
  title={Auctions with llm summaries},
  author={Dubey, Avinava and Feng, Zhe and Kidambi, Rahul and Mehta, Aranyak and Wang, Di},
  booktitle={Proceedings of the 30th ACM SIGKDD Conference on Knowledge Discovery and Data Mining},
  pages={713--722},
  year={2024}
}

@article{takenami2025anchoring,
  title={How does cognitive bias affect large language models? a case study on the anchoring effect in price negotiation simulations},
  author={Takenami, Yoshiki and Huang, Yin Jou and Murawaki, Yugo and Chu, Chenhui},
  journal={arXiv preprint arXiv:2508.21137},
  year={2025}
}

@inproceedings{zheng2026marketbench,
  title={Market-Bench: Benchmarking Large Language Models on Economic and Trade Competition},
  author={Zheng, Yushuo and Duan, Huiyu and Zhang, Zicheng and Zhu, Yucheng and Min, Xiongkuo and Zhai, Guangtao},
  booktitle={Proceedings of the 64th Annual Meeting of the Association for Computational Linguistics (Volume 1: Long Papers)},
  pages={39893--39906},
  year={2026}
}

@article{taneva2019,
  title={Information design},
  author={Taneva, Ina},
  journal={American Economic Journal: Microeconomics},
  volume={11},
  number={4},
  pages={151--185},
  year={2019},
  publisher={American Economic Association 2014 Broadway, Suite 305, Nashville, TN 37203-2425}
}

@article{candogan2022networks,
  title={Persuasion in networks: Public signals and cores},
  author={Candogan, Ozan},
  journal={Operations Research},
  volume={70},
  number={4},
  pages={2264--2298},
  year={2022},
  publisher={INFORMS}
}

@article{candogan2023private,
  title={Optimal disclosure of information to privately informed agents},
  author={Candogan, Ozan and Strack, Philipp},
  journal={Theoretical Economics},
  volume={18},
  number={3},
  pages={1225--1269},
  year={2023},
  publisher={Wiley Online Library}
}

@inproceedings{feng2024rationality,
  title={Rationality-robust information design: Bayesian persuasion under quantal response},
  author={Feng, Yiding and Ho, Chien-Ju and Tang, Wei},
  booktitle={Proceedings of the 2024 annual ACM-SIAM symposium on discrete algorithms (SODA)},
  pages={501--546},
  year={2024},
  organization={SIAM}
}

@article{huang2025crmarena,
  title={Crmarena-pro: Holistic assessment of llm agents across diverse business scenarios and interactions},
  author={Huang, Kung-Hsiang and Prabhakar, Akshara and Thorat, Onkar and Agarwal, Divyansh and Choubey, Prafulla Kumar and Mao, Yixin and Savarese, Silvio and Xiong, Caiming and Wu, Chien-Sheng},
  journal={arXiv preprint arXiv:2505.18878},
  year={2025}
}

@article{vanhoek2022walmart,
  title   = {How {Walmart} Automated Supplier Negotiations},
  author  = {Van Hoek, Remko and DeWitt, Michael and Lacity, Mary and Johnson, Travis},
  journal = {Harvard Business Review},
  year    = {2022},
  month   = nov,
  url     = {https://hbr.org/2022/11/how-walmart-automated-supplier-negotiations}
}

@InProceedings{zhuge2024gptswarm,
  title = 	 {{GPTS}warm: Language Agents as Optimizable Graphs},
  author =       {Zhuge, Mingchen and Wang, Wenyi and Kirsch, Louis and Faccio, Francesco and Khizbullin, Dmitrii and Schmidhuber, J\"{u}rgen},
  booktitle = 	 {Proceedings of the 41st International Conference on Machine Learning},
  pages = 	 {62743--62767},
  year = 	 {2024},
  editor = 	 {Salakhutdinov, Ruslan and Kolter, Zico and Heller, Katherine and Weller, Adrian and Oliver, Nuria and Scarlett, Jonathan and Berkenkamp, Felix},
  volume = 	 {235},
  series = 	 {Proceedings of Machine Learning Research},
  month = 	 {21--27 Jul},
  publisher =    {PMLR},
  url = 	 {https://proceedings.mlr.press/v235/zhuge24a.html},
}

@article{zhang2025gdesigner,
  title={G-designer: Architecting multi-agent communication topologies via graph neural networks},
  author={Zhang, Guibin and Yue, Yanwei and Sun, Xiangguo and Wan, Guancheng and Yu, Miao and Fang, Junfeng and Wang, Kun and Chen, Tianlong and Cheng, Dawei},
  journal={arXiv preprint arXiv:2410.11782},
  year={2024}
}

@article{saadfalcon2025shrinking,
  title={Shrinking the generation-verification gap with weak verifiers},
  author={Saad-Falcon, Jon and Buchanan, E Kelly and Chen, Mayee F and Huang, Tzu-Heng and McLaughlin, Brendan and Bhathal, Tanvir and Zhu, Shang and Athiwaratkun, Ben and Sala, Frederic and Linderman, Scott and others},
  journal={arXiv preprint arXiv:2506.18203},
  year={2025}
}

@inproceedings{zhou2026mass,
  title={Multi-agent design: Optimizing agents with better prompts and topologies},
  author={Zhou, Han and Wan, Xingchen and Sun, Ruoxi and Palangi, Hamid and Iqbal, Shariq and Vuli{\'c}, Ivan and Korhonen, Anna and Arik, Sercan},
  booktitle={International Conference on Learning Representations},
  volume={2026},
  pages={15844--15872},
  year={2026}
}

@article{alzu2026roma,
  title={ROMA: Recursive Open Meta-Agent Framework for Long-Horizon Multi-Agent Systems},
  author={Alzu'bi, Salaheddin and Nama, Baran and Kaz, Arda and Eswaran, Anushri and Chen, Weiyuan and Khetan, Sarvesh and Bala, Rishab and Vu, Tu and Oh, Sewoong},
  journal={arXiv preprint arXiv:2602.01848},
  year={2026}
}

@techreport{lloyds2025cdi,
  author      = {{Lloyds Banking Group}},
  title       = {2025 {Consumer Digital Index}: The {Generative AI} Edition},
  institution = {Lloyds Banking Group},
  year        = {2025},
  month       = nov,
  url         = {https://www.lloydsbankinggroup.com/assets/pdfs/media/consumer-digital-index/2025/2025-consumer-digital-index.pdf}
}

@article{bansal2025magentic,
  title={Magentic marketplace: An open-source environment for studying agentic markets},
  author={Bansal, Gagan and Hua, Wenyue and Huang, Zezhou and Fourney, Adam and Swearngin, Amanda and Epperson, Will and Payne, Tyler and Hofman, Jake M and Lucier, Brendan and Singh, Chinmay and others},
  journal={arXiv preprint arXiv:2510.25779},
  year={2025}
}

@article{myerson1983efficient,
  title={Efficient mechanisms for bilateral trading},
  author={Myerson, Roger B and Satterthwaite, Mark A},
  journal={Journal of economic theory},
  volume={29},
  number={2},
  pages={265--281},
  year={1983},
  publisher={Elsevier}
}

@article{nash1950equilibrium,
  title={Equilibrium points in n-person games},
  author={Nash, John F},
  journal={Proceedings of the national academy of sciences},
  volume={36},
  number={1},
  pages={48--49},
  year={1950},
  publisher={USA}
}
\bibliographystyle{iclr2027_conference}
\endgroup

\clearpage
\appendix
\etocdepthtag.toc{appendix}
\begingroup
\fontsize{10}{12}\selectfont
\setlength{\parskip}{1pt}
\etocsettagdepth{main}{none}
\etocsettagdepth{appendix}{subsection}
\etocsettocstyle{%
  \section*{Appendix Contents}
  \vspace{-3pt}
  \noindent\textcolor{RuleGray}{\rule{\linewidth}{0.4pt}}\par
  \vspace{3pt}
}{}
\etocsetstyle{section}{}{}{%
  \par\addvspace{7pt}\noindent
  \makebox[2em][l]{\textbf{\etocnumber}}\textbf{\etocname}%
  \nobreak\hfill\makebox[2em][r]{\textbf{\etocpage}}\par
}{}
\etocsetstyle{subsection}{}{}{%
  \noindent\hspace*{2em}\makebox[2.8em][l]{\etocnumber}\etocname
  \nobreak\textcolor{RuleGray}{\dotfill}\makebox[2em][r]{\etocpage}\par
}{}
\sbox{\AppendixContentsBox}{%
  \begin{minipage}{\textwidth}
    \tableofcontents
  \end{minipage}%
}
\noindent
\ifdim\dimexpr\ht\AppendixContentsBox+\dp\AppendixContentsBox\relax>0.95\textheight
  \resizebox*{!}{0.95\textheight}{\usebox{\AppendixContentsBox}}%
\else
  \usebox{\AppendixContentsBox}%
\fi
\par
\endgroup
\clearpage
\normalsize

\section{Extended Related Work}
\label{app:extended-related}
\paragraph{Negotiation and markets with LLM agents.}
Language-based negotiation research involves discourse, planning, and recurrent interaction \citep{lewis2017,he2018,akata2023}. NegotiationArena and GLEE provide verifiable outcome games, and measures of individual gains, efficiency, and fairness \citep{bianchi2024negotiationarena,shapira2024glee}. TERMS-Bench also evaluate agents' negotiation skills \citep{zhang2026termsbench}. Negotiated prices are related to bargaining theory in studies of information asymmetry and patience \citep{deng2024bargaining}. Experiments on buyer--seller differences \citep{xia2024bargaining} and opening demands \citep{takenami2025anchoring} reveal the effect of strategic position on outcomes. Together these research motivate evaluating both whether agents agree and who benefits more. 

The same questions arise in marketplaces with multiple participants. AucArena evaluates bidding, planning and budget management \citep{chen2023aucarena}, and Market-Bench examines procurement and retail competition \citep{zheng2026marketbench}. Pricing experiments demonstrate collusion and sensitivity to prompt changes \citep{fish2024collusion} and Project Vend and Anthropic's Project Deal explore agents acting in commercial settings \citep{anthropic2025vend,anthropic2026deal}. Literature complementary develops allocation and payment rules for LLM-generated material \citep{duetting2024mechanism, dubey2024summaries}. Our experiments address these restrictions and vary disclosure, comparing proposals with final outcomes across models. This disentangles information access as a design decision in the interaction.

\paragraph{Information design and strategic behavior.}
Classical studies of bargaining and markets focus on the role of individual incentives in determining agreement and the division of gains \citep{nash1950,rubinstein1982,chatterjee1983,smith1962}. Information allows for a further differentiation: it is useful for an optimal individual decision-maker who can ignore it \citep{blackwell1953} but privately valuable information need not be welfare improving \citep{hirshleifer1971}. Bayesian persuasion and Bayes correlated equilibrium \citep{kamenica2011,bergemann2016,bergemann2019} provide tools for designing disclosure around agents' decisions and incentives. Conflicts of interest can limit communication \citep{crawford1982}, public signals can be overweighted in coordination \citep{morris2002}, and partial verifiable disclosure can reduce inefficient screening \citep{glode2018}. 

Information design is concerned with how private signals jointly affect decisions when multiple agents interact \citep{taneva2019}. Extensions consider interactions in networks \citep{candogan2022networks} and disclosure to agents with prior private information \citep{candogan2023private}. Work on bounded rationality develops policies robust to uncertainty about how receivers respond \citep{feng2024rationality}. These contributions show why disclosure must be evaluated in relation to the interaction and the designer's objective. We apply this perspective empirically to LLM agents, measuring their responses to truthful information without prescribing an equilibrium strategy.

\paragraph{Multi-agent system design.} CAMEL and AutoGen make roles and conversations configurable components of agent systems \citep{li2023camel,wu2023autogen}. GPTSwarm optimizes prompts and connections between agents \citep{zhuge2024gptswarm}, G-Designer learns task-dependent communication graphs \citep{zhang2025gdesigner}, and Multi-Agent System Search explores prompts and interaction structures \citep{zhou2026mass}. We examine the complementary choice of what information agents receive within fixed roles and interaction rules. 

Studies of asymmetric knowledge, collective information processing, conformity, and failures to share private evidence further show why individual capabilities do not guarantee effective coordination \citep{yashwanth2026sotopiatom,riedl2026,anthropic2026multiagent}. Oversight studies provide a related perspective on information use. Selecting relevant trajectory excerpts can improve sabotage detection \citep{arike2026}, and distributing evaluation requirements across judge calls can improve agreement with experts under matched evidence and budgets \citep{akinwande2026}. These approaches change how evidence is processed. Our settings add a strategic consideration, since accurate information can help an agent pursue its own gains while changing its demands in ways that reduce collective performance. \looseness=-2

\clearpage
\section{Multi-Resource Trade}
\label{app:trade-results}

This appendix explains how we construct the trading markets and interpret the additional results. We first describe the two-resource study, which varies whether A or B has more opportunity to increase its own gain at the expense of total gains. We then examine the proposed trades, and each agent's final gains. The remaining studies vary conversation length, resource count, and the use of a mediator.

\FloatBarrier
\Needspace{14\baselineskip}
\subsection{Market construction and validation}
\label{app:trade-methods}

\subsubsection{Markets and interaction protocol}

A sends the first message, and the agents then alternate turns. They can propose a trade, accept the other's standing offer, or send a message without proposing. A counteroffer replaces the standing offer. Agents can ask about preferences and discuss quantities throughout the conversation. Each call includes both inventories, the current offer, and the complete public conversation. Responses use the JSON format described in \S\ref{app:trade-prompts}. The main study allows eight messages in total and at most three proposals per agent. More generally, an even message budget $T$ permits up to $\max(1,T/2-1)$ proposals per agent. The final message allows a response to the standing offer but no new proposal, because the counter-party would have no turn left to answer. Acceptance ends the negotiation; otherwise, reaching the message limit leaves both agents with their initial inventories.

Each market has two resources. A holds $q$ units of one and one unit of the other, while B has the opposite inventory, with $q\in\{2,3,4,5,6\}$. Each agent values the resource abundant in the other's inventory more highly. The value pairs in the table list an agent's lower and higher per-unit values. For example, $(5,95)$ represents a strong preference, while $(45,55)$ represents a mild preference. Both pairs sum to $100$, so the difference concerns preference strength rather than the scale of an agent's values. For each row and inventory quantity, we assign the two profiles to A and B, reverse the assignment, and give both agents the first profile. This produces $6\times5\times3=90$ markets. Resource names are balanced across the construction. Each model faces these equivalent markets under least, coarse, and full information.

\begin{table}[H]
\centering
\caption{\textbf{Preference strengths used to construct the two-resource markets.} Values are per unit of the less- and more-preferred resources. For each row and inventory quantity, we assign the first pair to A and the second to B, reverse the assignments, or give both agents the first pair.}
\small
\renewcommand{\arraystretch}{1.18}
\setlength{\tabcolsep}{10pt}
\begin{tabularx}{\linewidth}{@{}c*{4}{>{\centering\arraybackslash}X}@{}}
\toprule
 & \multicolumn{2}{c}{\textbf{First value pair}} & \multicolumn{2}{c}{\textbf{Second value pair}} \\
\cmidrule(lr){2-3}\cmidrule(l){4-5}
\textbf{Profile} & Less-preferred & More-preferred & Less-preferred & More-preferred \\
\midrule
$1$ & $5$  & $95$ & $45$ & $55$ \\
\rowcolor{black!4} $2$ & $10$ & $90$ & $40$ & $60$ \\
$3$ & $20$ & $80$ & $30$ & $70$ \\
\rowcolor{black!4} $4$ & $30$ & $70$ & $20$ & $80$ \\
$5$ & $40$ & $60$ & $10$ & $90$ \\
\rowcolor{black!4} $6$ & $45$ & $55$ & $5$ & $95$ \\
\bottomrule
\end{tabularx}
\end{table}

\subsubsection{Which agent has greater extraction opportunity?}

An extraction opportunity allows an agent to increase its own gain while reducing total gains and still benefiting its counter-party agent. Using the fixed inventories and values, we compare each agent's best gain from such a trade with its gain under \emph{full exchange}, where both agents transfer all units of their abundant resource. We consider integer transfers in these directions that give both agents positive gains. The agent with the larger possible increase has greater extraction opportunity; equal increases define the equal-opportunity group. This calculation yields $30$ markets per group and is performed before inspecting model responses. Profile order alone does not determine the group, because relative preference strengths vary across table rows. Group labels are not shown to agents, and A starts negotiating. \looseness=-2

\subsubsection{Valid actions and extractive counteroffers}

The game checks that proposed trades respect inventories and that agents accept only the counter-party's standing offer. Invalid actions do not execute, but remain recorded in the conversation. We retain all negotiations in the reported results. As a robustness check, we exclude markets with an invalid action under either disclosure condition and find that the qualitative conclusions remain unchanged. We also measure whether agents use their opportunities through \emph{extractive counteroffers}. An extractive counteroffer increases its proposing agent’s gain while reducing total gains relative to the counter-party’s preceding valid offer, as calculated from the proposed quantities and true values. Its frequency also depends on how many opportunities agents have to counteroffer.

\subsection{Individual gains across information conditions}
Figure~\ref{fig:trade-surplus-app} complements the main-text comparison of full versus coarse disclosure by showing each agent's realized gains under all three information conditions. Each gain is normalized by the market's maximum total gain, so A's and B's contributions sum to realized performance $W$ where disagreements contribute zero. Each opportunity group contains $30$ markets that are repeated across information conditions. The corresponding candidate quality and realization gaps appear in Figure~\ref{fig:cross-arena-market-groups}. \looseness=-2

Under full disclosure, Claude Sonnet 5 and Grok 4.3 increase both agents' gains in every opportunity group. GPT-6 Sol, Claude Opus 5.5, and Grok 4.7 instead shift gains toward the agent with greater extraction opportunity while reducing the counterpart's gains by more. These differences distinguish improvements in total gains from changes in their distribution.

\begin{figure}[!htbp]
\centering
\includegraphics[width=\textwidth]{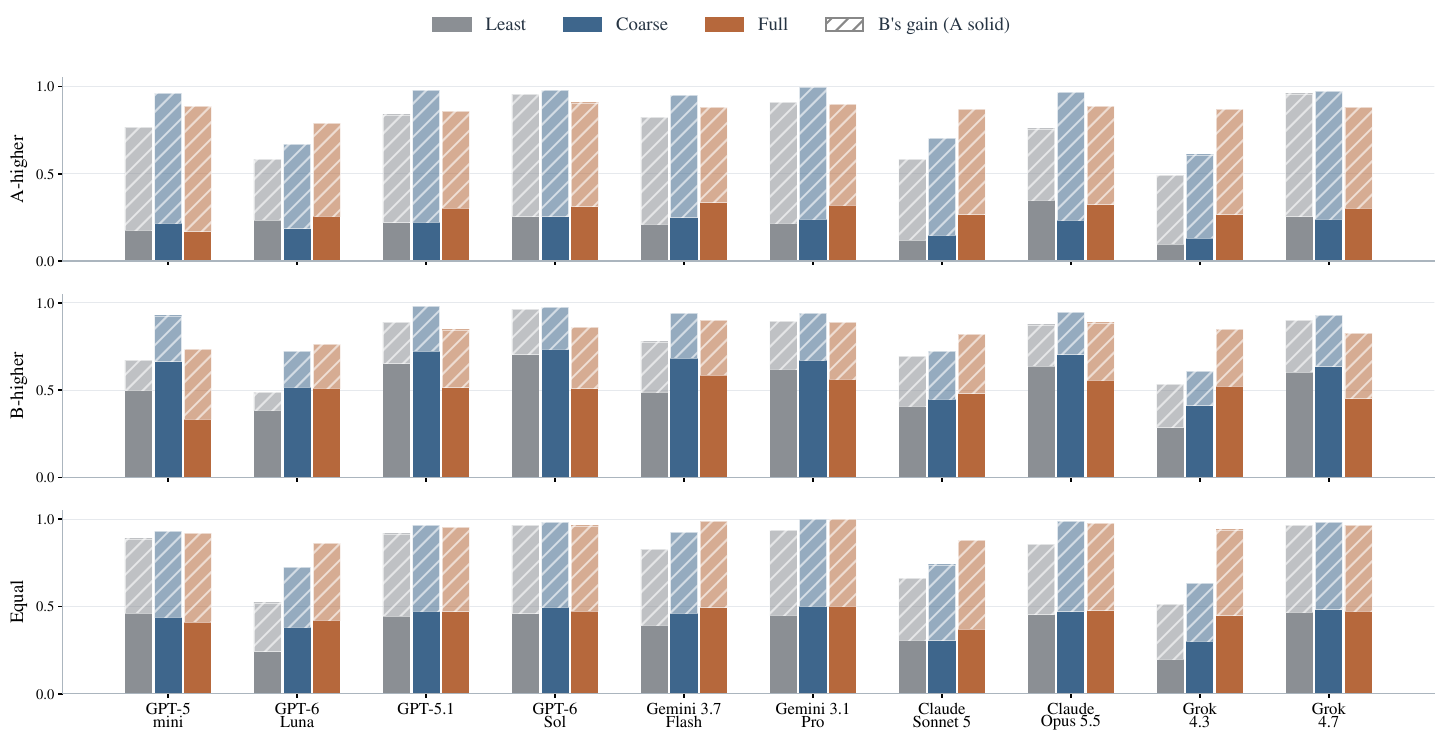}
\caption{\textbf{Individual gains by extraction-opportunity group.} Solid segments show A's normalized gain and hatched segments B's; total heights give realized performance $W$. Gray, blue, and orange denote least, coarse, and full information.}
\label{fig:trade-surplus-app}
\end{figure}

\subsection{Dialogue length and resource count}

These studies examine when dialogue can compensate for limited initial information and when more complex trades make additional detail more useful. A longer conversation gives agents time to learn about each other's preferences. More resources require them to compare a wider range of possible exchanges. Both studies focus on markets where B has greater extraction opportunity, rather than covering all three groups we used in the main study.

\subsubsection{More time to negotiate}

The dialogue study uses six markets from the $30$-market panel where B has greater extraction opportunity, combining three preference profiles with abundant-resource quantities of two or three units. We use the same markets at message limits of two, four, six, eight, and twelve. A moves first, and proposal allowances increase with the message budget according to the rule above. Extending dialogue from two to twelve messages improves welfare more under least than full information for eight of the ten models, with GPT-6 Sol and Grok 4.7 as exceptions. Intermediate changes remain uneven (Figure~\ref{fig:trade-horizons-app}(b)). This suggests that conversation can help compensate for limited initial information.

\subsubsection{Trading with more resources}
\label{app:resource-construction}

We study markets with two, six, and ten resources. We begin with six markets containing two resources, X and Y, with different inventories and values. Each market is also tested with resource names swapped, giving $12$ markets at each size. A initially owns more X, while B owns more Y. To reach six resource types, we add four new resources. A owns more units than B of two new resources, and B owns more units than A of the other two. We repeat this process to reach ten resource types. For each resource, one agent owns one unit and the other owns between two and six units. Adding resources expands the inventories without changing the quantities or values of resources already included. Utilities add across resources, and preference rankings have no ties. Agents can exchange any available quantities, including several resource types for a single type. \looseness=-2

The added resources introduce different trading decisions. With two resources, B can increase its own gain while reducing total gains and still benefiting A. With six resources, a proposal can benefit both agents by combining transfers that favor different sides. With ten resources, agents must also choose which resources to include, because some transfers reduce total gains. Exchanging all units of each agent's more abundant resources maximizes total gains at six resources, but achieves only $43$--$72\%$ of the maximum at ten. At every size, maximum total gains can be achieved while benefiting both agents. Adding resources changes both the available trades and the difficulty of finding them; extraction opportunity is not held constant. 

Confidence intervals resample the six underlying markets, keeping their resource-name versions together. At ten resources, full information improves candidate quality $Q$ for all ten models and realized performance $W$ for nine, with GPT-6 Sol nearly unchanged. Figure~\ref{fig:trade-sweeps}(a) decomposes these differences into changes in candidate quality and the realization gap. \looseness=-2

\begin{figure}[tbp]
  \centering
  \includegraphics[width=\textwidth]
    {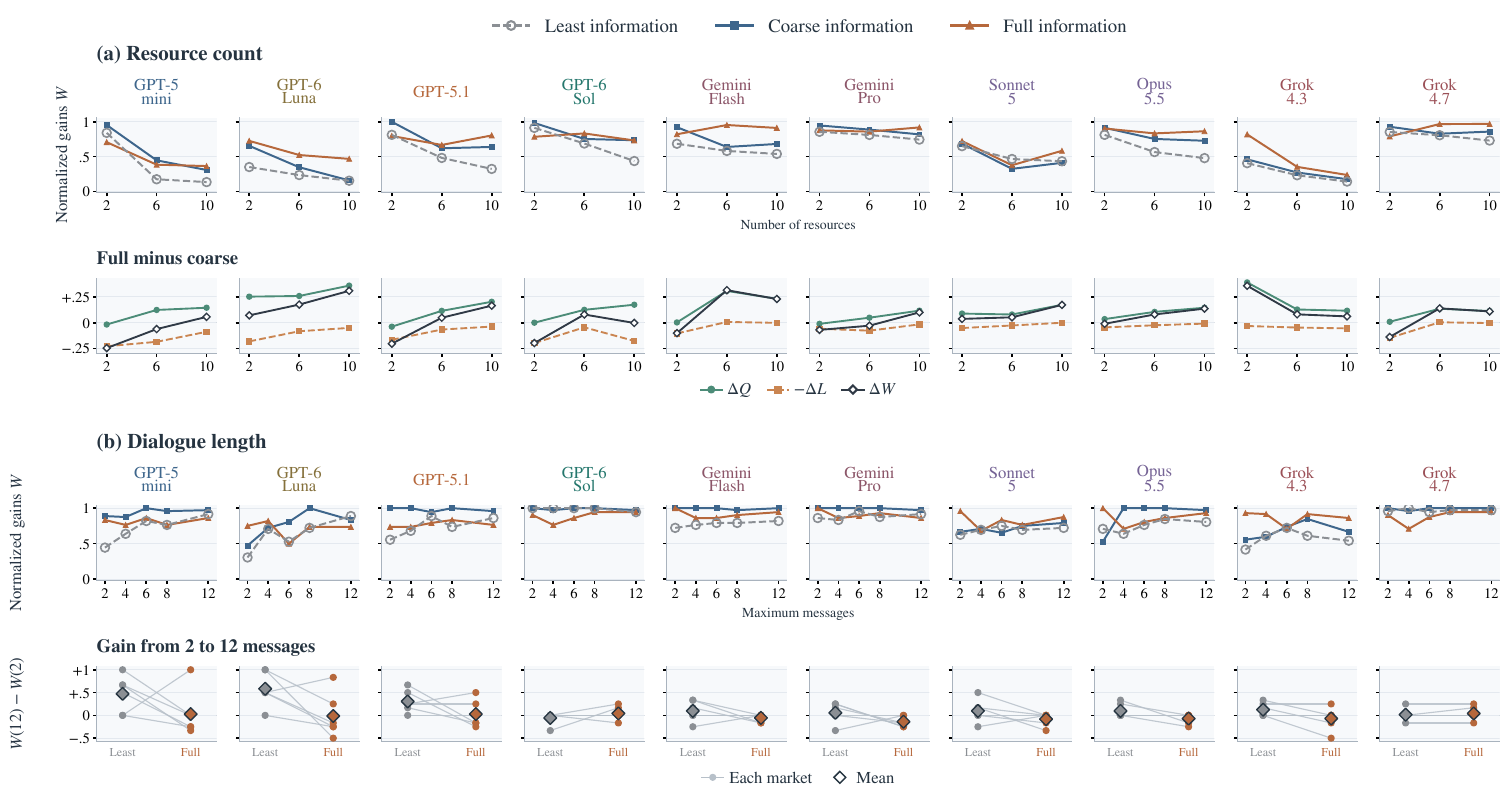}
  \caption{\textbf{Complete resource-count and dialogue-length studies.} (a) Realized performance at each resource count and the full-minus-coarse decomposition into changes in candidate quality, realization gap, and realized performance. (b) Realized performance across message limits and changes from two to twelve messages under least and full information. Faint lines connect the same markets; diamonds show means. Both studies use markets where B has greater extraction opportunity.}
  \label{fig:trade-design-studies-complete}
\end{figure}

\subsection{Welfare-oriented mediation}

Figure~\ref{fig:trade-mediator} compares direct negotiation with mediated trade on the same $30$ two-resource markets where B has more extraction opportunity. In direct negotiation, A and B seek their own gains and alternate messages for up to eight turns. In mediated trade, a third instance of the same model is instructed to maximize their total gains. It proposes a trade, and each negotiating agent privately accepts or rejects it. The trade occurs only if both agents accept the given proposal. These focused markets use the same inventories and utilities across the two protocols as we used in the main experiments as in Figure \ref{fig:cross-arena-accounting}. The comparison changes the decision-maker's objective, prompt, and communication protocol alongside the recipient of information. \looseness=-2

\begin{figure}[!h]
  \centering
  \includegraphics[width=0.9\textwidth]{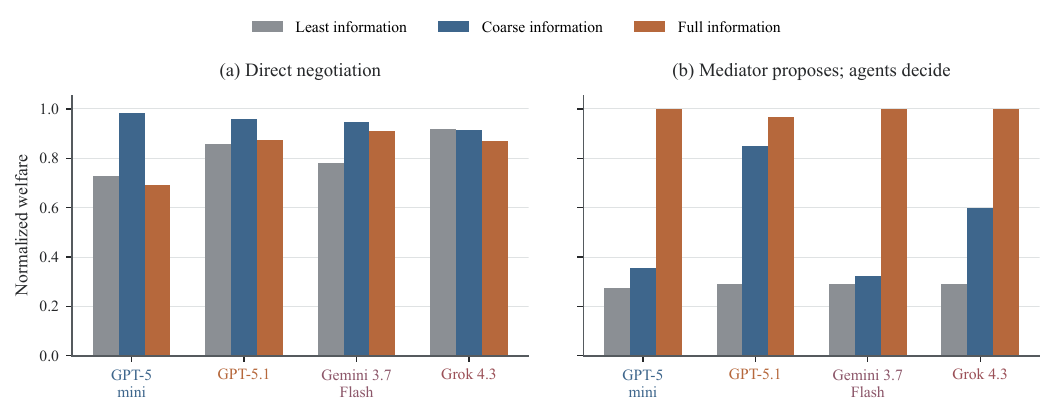}
  \caption{\textbf{Direct negotiation and mediator proposals.} Mean normalized welfare across the same $30$ markets per model and information condition, where B has greater extraction opportunity. In panel (a), agents negotiate directly; in panel (b), a mediator proposes a trade that both agents must accept. The mediator receives the value information specified by each disclosure condition.}
  \label{fig:trade-mediator}
\end{figure}

The preferred disclosure level is reversed between the two protocols. In these peer runs, coarse information produces higher mean welfare than full information for all four models. Instead, the mediator performs best when given complete information, achieving results that are close to the maximum overall gain. Direct negotiation produces higher welfare than mediation when there is little or coarse information: agents can exchange proposals and learn through dialogue, whereas the mediator must construct its proposal based on the information available at the time. The findings demonstrate how precise values can be useful in developing a high-welfare proposal, even if disclosing them to self-interested negotiators undermines the final agreement. They advocate comparing entire interaction designs rather than focusing solely on the causal effect of information placement.

\newpage

\section{Single-Good Bargaining}
\label{app:bargaining-results}

Bargaining separates the creation of surplus through agreement from its
division through the negotiated price. We examine both outcomes under
symmetric and asymmetric disclosure, distinguishing the identity of the first
mover and the time available for counteroffers.
Candidate quality, realized performance, and realization gaps at six messages are reported by surplus group in Figure~\ref{fig:cross-arena-market-groups}.

\FloatBarrier
\Needspace{14\baselineskip}
\subsection{Market construction and validation}
\label{app:bargaining-methods}

We draw costs and values independently from the stated $U[20,100]$ prior, then discard pairs with $v\le c$ and retain profitable pairs to fill three surplus groups: thin ($0<v-c\le15$), medium ($15<v-c\le35$), and large ($v-c>35$). There are five base pairs in each group, each played with both the seller and the buyer moving first. Agents are told that buyer values and seller costs are drawn independently and uniformly from $20$ to $100$. Agents receive the unconditional prior and are not told the market-selection procedure or surplus group. Coarse disclosure nevertheless reveals whether trade can benefit both sides, and full disclosure reveals the exact utility values. We average results across $30$ negotiations: ten each with thin, medium, and large surplus. Markets where trade cannot benefit both sides are excluded.

Agents may propose or counteroffer only if the other agent has a turn left to respond. On the final turn, they can only accept or reject the current offer. With a two-message limit, the first agent proposes a price and the second accepts or rejects it. Instructions describing the agent's role, the market background, and its objective remain the same across information conditions. Only the disclosed information changes. The prompt and information cards appear in \S\ref{app:bargaining-prompts}.

\subsection{Information recipient and starting role}

\begin{figure}[!h]
\centering
\includegraphics[width=\textwidth]{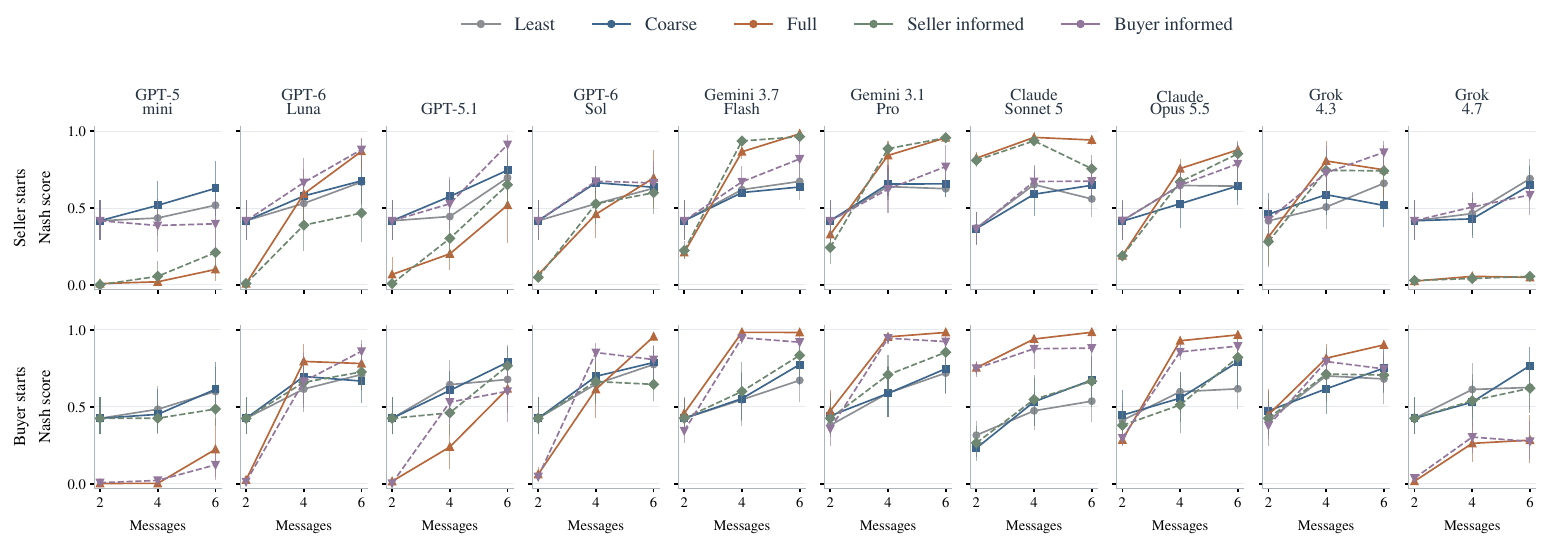}
\caption{\textbf{Normalized Nash scores by information condition and starting role.} Rows separate seller-first and buyer-first negotiations; columns show models. Each point averages $15$ markets at the indicated message limit. Equal division scores one; unequal agreements score less, and disagreement scores zero. Error bars show $95\%$ bootstrap confidence intervals for the mean, obtained by resampling cost--value pairs.}

\label{fig:bargaining-roles-nash}
\end{figure}

We compare disclosure conditions by starting role and message limit to examine how information and first-mover advantage interact. Figure~\ref{fig:bargaining-roles-nash} reports normalized Nash scores, while Figure~\ref{fig:bargaining-role-gains} shows each agent's gain as a fraction of available surplus. One-sided disclosure gives the informed agent exact information and its counter-party least information, so comparisons with coarse disclosure change both agents' information. With two messages and the seller moving first, full and seller-only disclosure increase the seller's mean normalized gain over coarse disclosure for all ten models. When buyers start, seller-only disclosure leaves sellers' gains unchanged or lower for nine of the ten models. Information therefore helps differently depending on who makes the first offer, with more model-dependent patterns in longer conversations. \looseness=-2

\newpage

\begin{figure}[!h]
\centering
\includegraphics[width=1\textwidth]{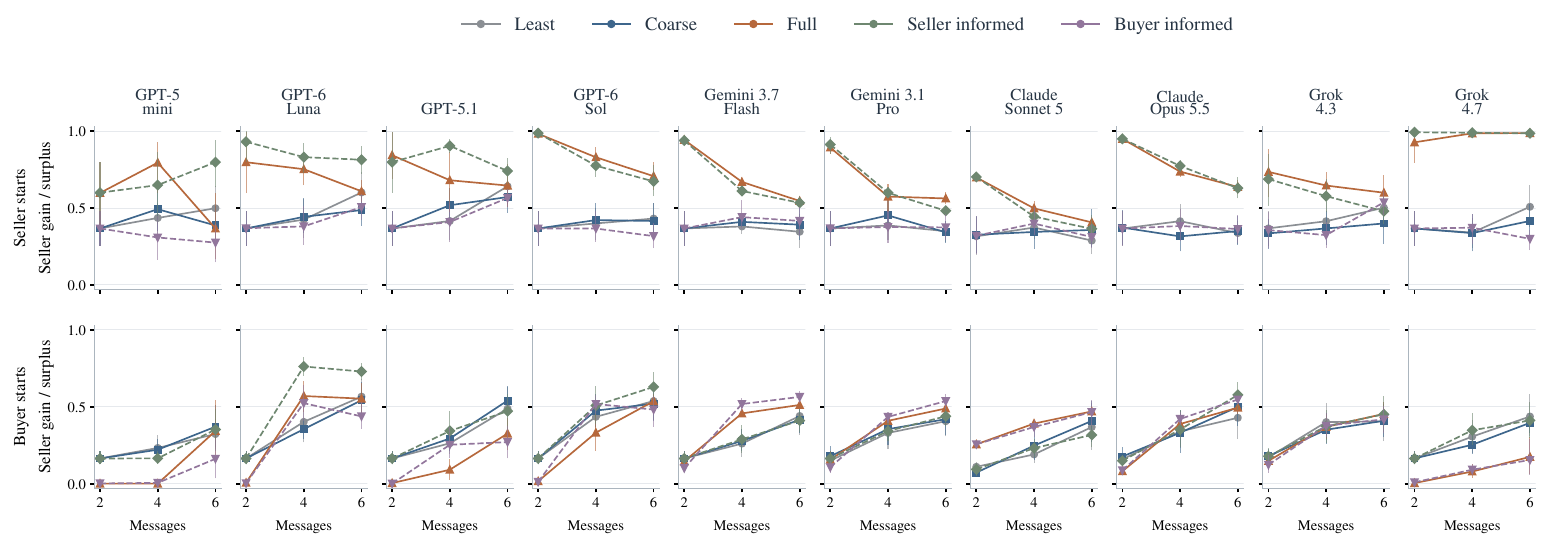}

\includegraphics[width=1\textwidth]{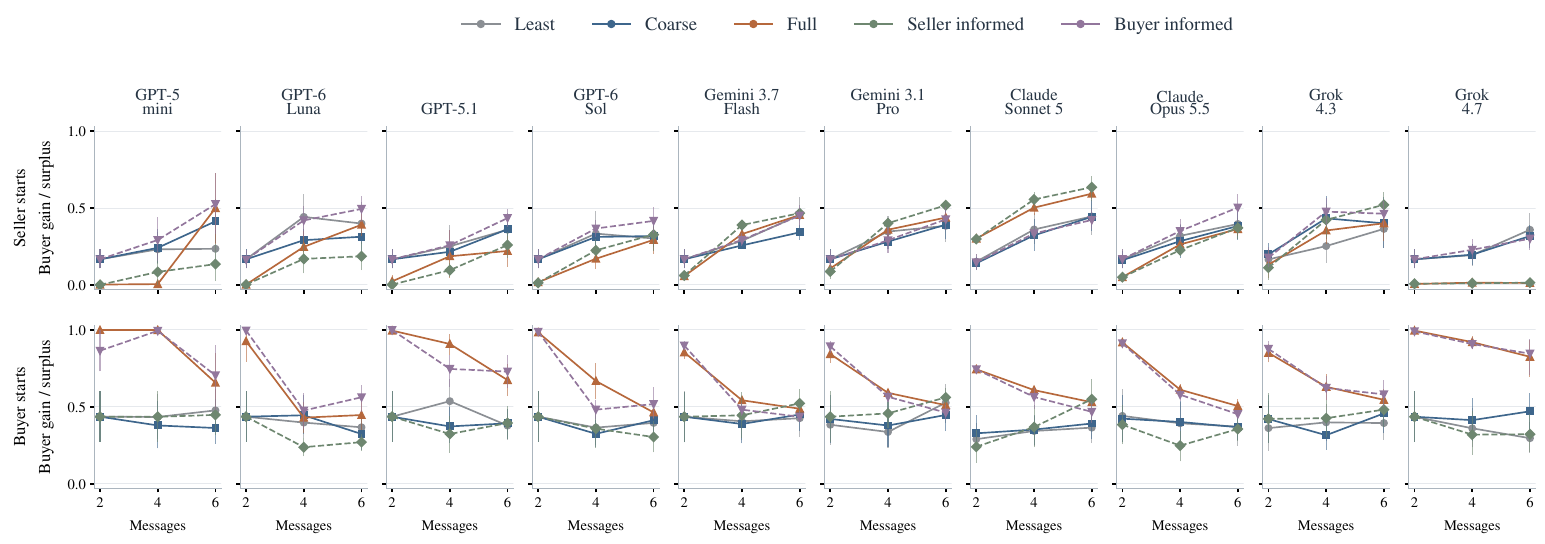}
\caption{Average gains earned by sellers and buyers. The upper block shows seller gains $(p-c)/(v-c)$; the lower block shows buyer gains $(v-p)/(v-c)$, with zero for disagreement. Rows separate starting roles and columns show models. Each point averages $15$ markets. Error bars show $95\%$ confidence intervals from bootstrap resampling of cost--value pairs.}
\label{fig:bargaining-role-gains}
\label{fig:bargaining-roles-seller-gain}
\label{fig:bargaining-roles-buyer-gain}
\end{figure}

\vspace{0.6cm}
\subsection{Surplus size}
Figure~\ref{fig:bargaining-strata} shows the full-minus-coarse change in realized performance by surplus group and message limit. Negative values favor coarse information. In large-surplus markets, full disclosure lowers scores for all four GPT models at every message limit, although longer dialogue reduces these losses. Grok 4.7 also shows persistent losses, whereas Claude Sonnet 5 benefits throughout. These differences are averaged together with thin- and medium-surplus markets in the main-text comparison.

\begin{figure}[!h]
  \centering
  \includegraphics[width=\textwidth]{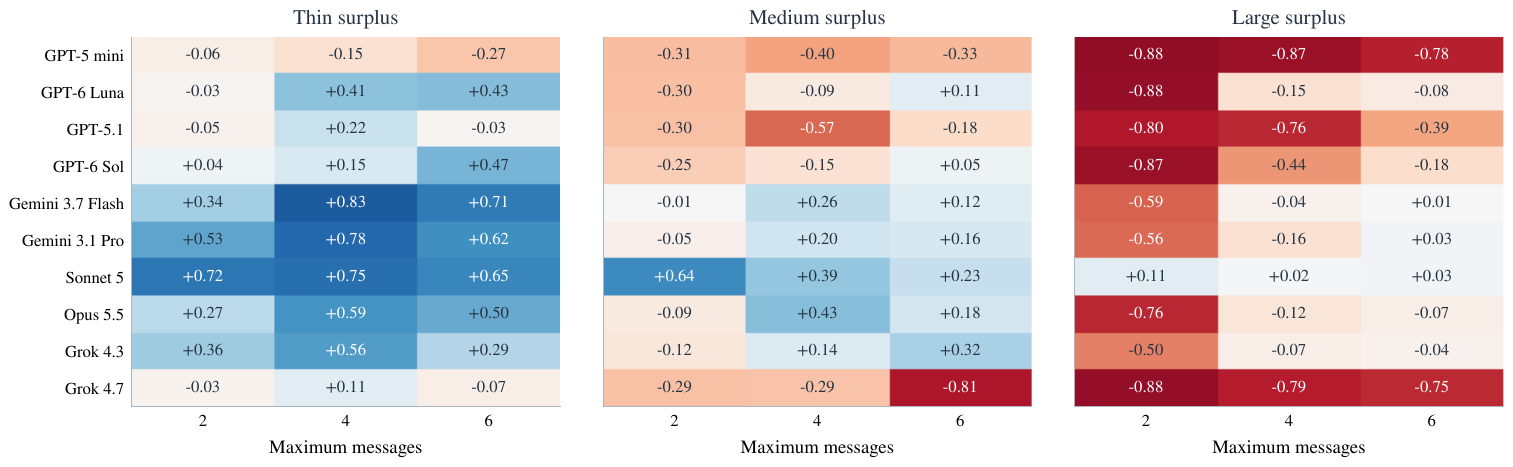}
  \caption{\textbf{Changes in bargaining performance by available surplus.} Cells show full-minus-coarse differences in mean normalized Nash scores, including zero for disagreements. Each surplus group contains five cost--value pairs, each evaluated under both starting roles.}
  \label{fig:bargaining-strata}
\end{figure}

\newpage
\section{Double Auctions}
\label{app:auction-results}

A buyer and seller can trade only if the buyer's bid is at least the seller's ask. We examine how disclosure changes these prices, the surplus that is cleared, and its division between buyers and sellers.

\subsection{Market construction and order protocol}
\label{app:auction-methods}

We construct markets that differ in how many trades can benefit both sides and how much gain remains on the last profitable trade. Each candidate market has three buyer values and three seller costs, drawn independently and uniformly from $0$ to $100$ and rounded to one decimal place. To identify the trades that maximize total gains, we pair the highest-value buyer with the lowest-cost seller, then the next-highest buyer with the next-lowest seller, and so on. 

We count pairs for which the buyer's value exceeds the seller's cost. This number, $q^*$, can be one, two, or three; we exclude markets with no profitable pair. The last profitable pair has the smallest positive gain among these pairs. We classify that gain as narrow ($[2,8]$), medium ($[12,24]$), or wide ($[35,50]$). Thus, a market can support three profitable trades while having a smaller gain on the third.

We combine each trade count with each gain range, giving nine market groups. Within each group, we select three or four markets spread across the candidate pool's gain distribution using evenly spaced quantiles, without changing the drawn values or costs. The resulting $30$ markets include ten with each trade count and ten in each gain range. This lets us vary the number of profitable trades separately from the gain on the last one. These quantities describe what is possible under the true values and costs and the agents' submitted prices determine how many trades actually occur.

Each agent knows its own value or cost and the identities of all three buyers
and three sellers. The prompt explains simultaneous submission, midpoint pricing, and the rule that buyers cannot bid above their values and sellers cannot ask below their costs. It does not reveal the generating distribution, the market's margin group, or the efficient number of trades. The agent is asked to maximize its own expected payoff and return a structured order (\S\ref{app:auction-prompts}). The gap between buyer value and seller cost determines the surplus available from a trade. Disclosure can change how much of that surplus each side seeks. We examine this behavior by varying information recipients and market size.

\subsection{Welfare and losses at market clearing}
We compare the gains available from buyer–seller pairs formed by submitted prices with the gains from completed trades. This separates changes in who is paired from gains lost when buyers offer less than sellers ask, using the same markets for both measures.

\begin{figure}[H]
  \centering
  \includegraphics[width=\textwidth]{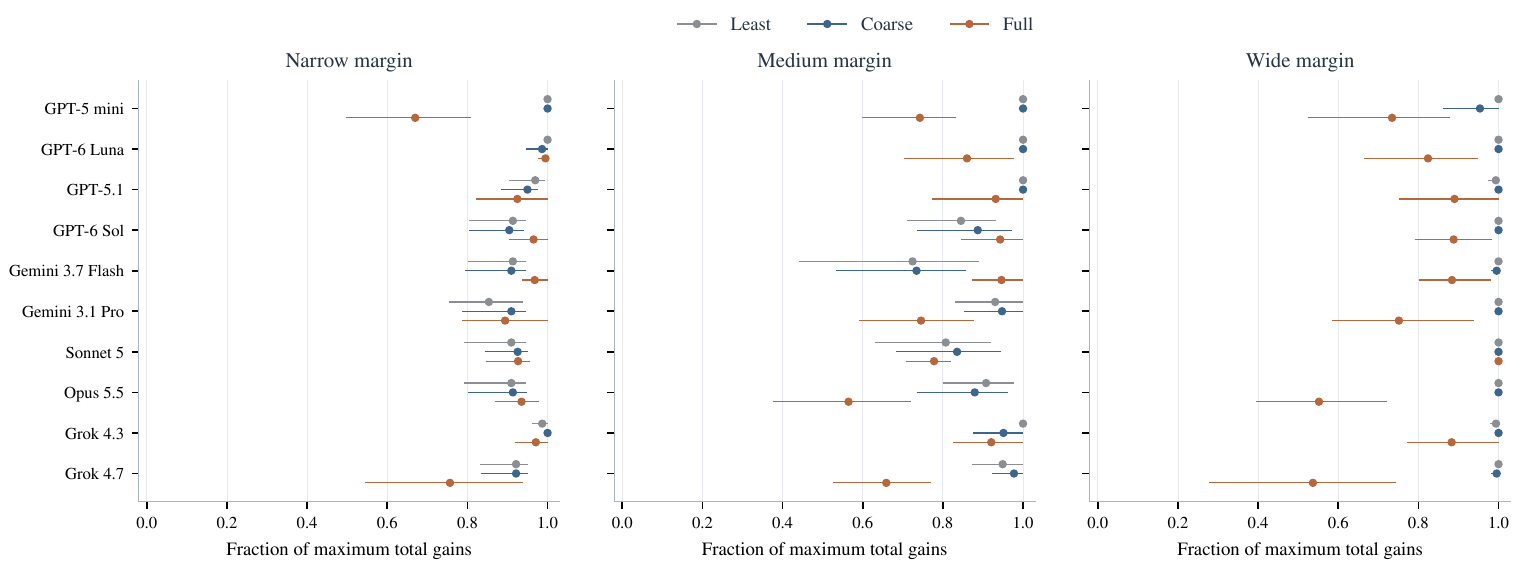}
  \vspace{-10pt}
  \caption{\textbf{Auction welfare by model, information level, and margin group.} Normalized welfare in ten narrow, ten medium, and ten wide margin markets per model. Margin groups describe the gain from the last efficient buyer--seller pair.}
  \label{fig:auction-families}
\end{figure}

Full disclosure lowers overall realized gains for nine of the ten models, with Gemini 3.7 Flash the exception (Figure~\ref{fig:auction-families}). The effects also vary across the margin groups. GPT-6 Sol benefits in narrow and medium margin markets from more information, but loses in wide-margin markets. At wide margins, full disclosure lowers gains for nine models, but leaves Claude Sonnet 5 unchanged, with the largest declines for Claude Opus 5.5 and Grok 4.7. Candidate quality remains above $0.99$ under both coarse and full disclosure for every model, so these losses primarily reflect profitable trades that do not clear.

\FloatBarrier
\Needspace{14\baselineskip}
\subsection{Asymmetric disclosure}
This ablation gives all agents coarse information including value and cost rankings, then adds exact values and costs for buyers only, sellers only, or both sides. It uses $15$ matched markets, with five each supporting one, two, or three efficient trades. Market construction balances these three groups, while the main disclosure experiment crosses efficient quantity with margin size.

Across all four tested models, giving exact information only to buyers lowers sellers' share relative to coarse disclosure, whereas giving it only to sellers raises their share (\cref{fig:auction-placement}). Full disclosure to both sides reduces total gains for the GPT models and Grok but improves Gemini's outcome. These shares describe the division of realized gains; a larger share can accompany lower total earnings.

\begin{figure}[!htbp]
  \centering
  \includegraphics[width=0.75\textwidth]{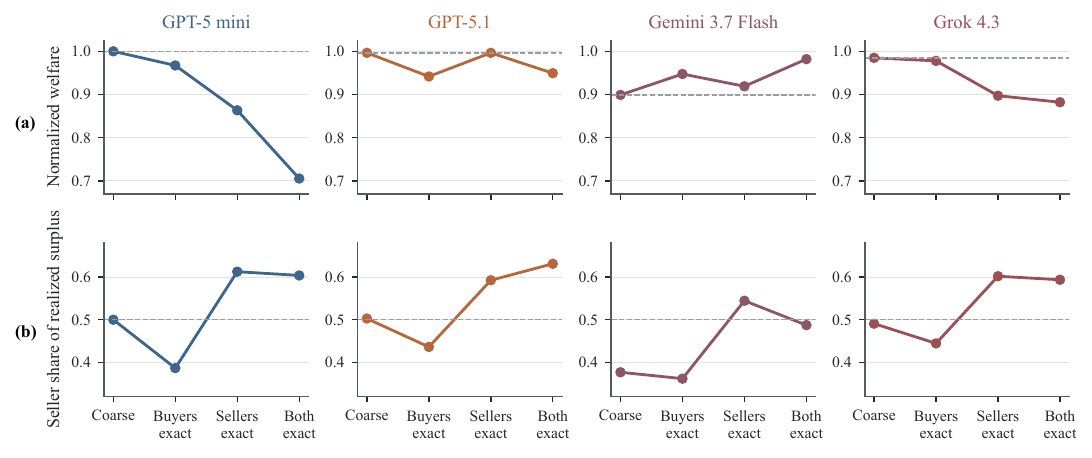}
  \caption{\textbf{Auction welfare and seller shares under asymmetric disclosure.} All agents receive coarse rankings; exact values and costs are additionally given to buyers only, sellers only, or both sides. The top row shows normalized welfare; the bottom row shows sellers' share of total realized gains. Dashed lines mark coarse-information welfare and equal division, respectively. Each model uses the same $15$ markets across conditions. \looseness=-2}
  \label{fig:auction-placement}
\end{figure}

\newpage

\section{Additional Results and Behavioral Analysis}

\subsection{Candidate quality and realized performance by market group}
\label{app:cross-arena-market-groups}

Figure~\ref{fig:cross-arena-market-groups} expands the overview in Figure~\ref{fig:cross-arena-accounting} by separating market groups in each setting. This shows whether disclosure changes the opportunities that agents identify, the outcomes they reach, and how these vary with market structure. Trade groups differ in extraction opportunity, bargaining groups in available surplus, and auction groups in the gain available to the last profitable buyer--seller pair. \looseness=-2

\begin{figure}[H]
  \centering
  \includegraphics[width=\textwidth]{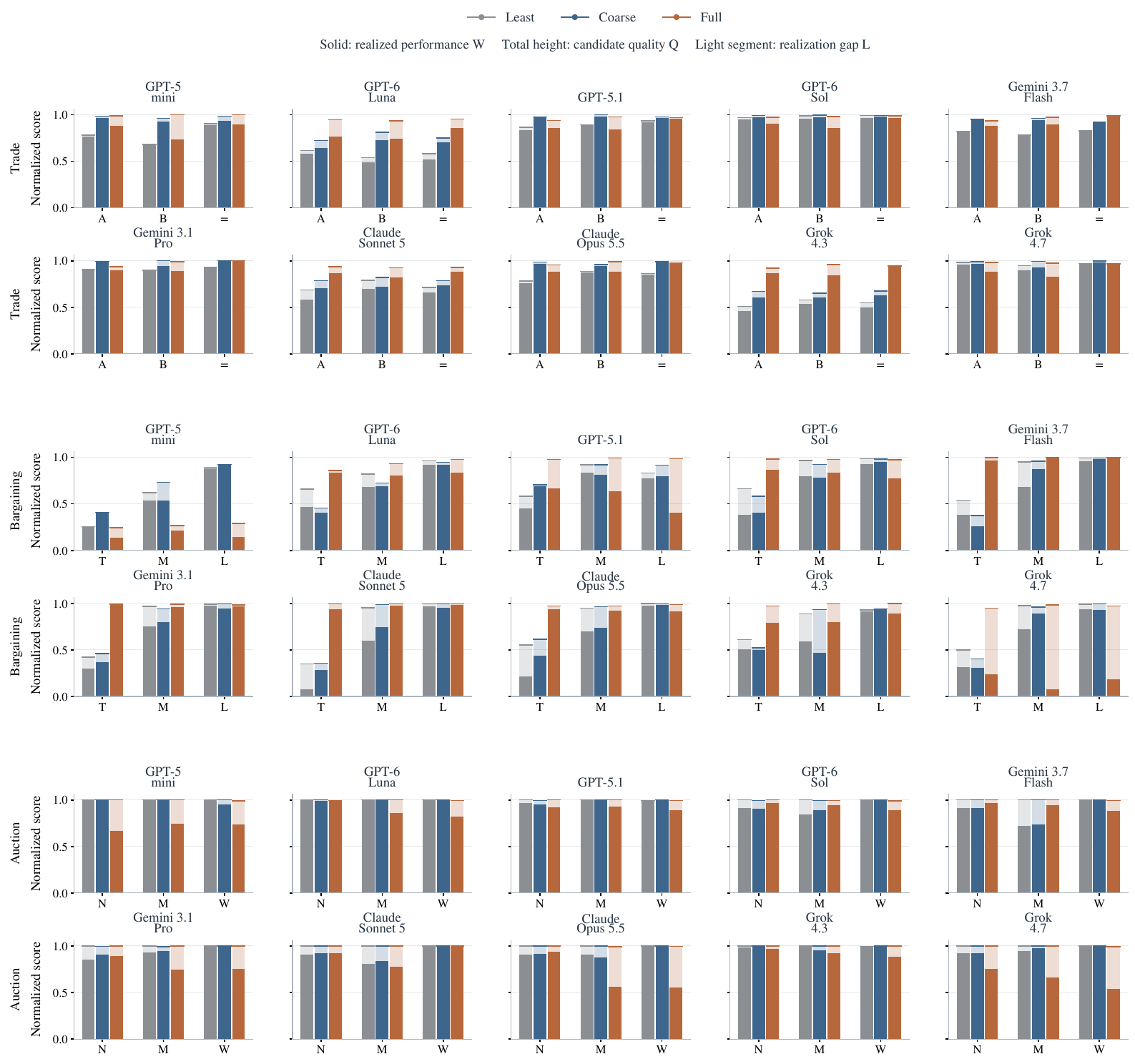}
  \caption{\textbf{Candidate quality, realized performance, and realization gaps by market group.} Columns show models; rows show the three settings. Within each, bars compare least, coarse, and full disclosure. Solid segments show realized performance $W$, total heights show candidate quality $Q$, and the remaining segments show the realization gap $L=Q-W$. Trade and auctions measure normalized total gains; bargaining measures the normalized Nash product, giving zero for disagreements. Each condition includes $30$ trade markets per group, $10$ bargaining instances per group at six messages across both starting roles, and $10$ auction markets per margin group. \looseness=-2}
  \label{fig:cross-arena-market-groups}
\end{figure}

\newpage
\subsection{Mixed-model Interactions}
\label{app:mixed-qwl}
\label{app:mixed-models}

This section expands the mixed-model ablation results in \cref{sec:mixed-models}. We test all six combinations of GPT-5.1, Claude Sonnet 5, Gemini 3.7 Flash, and Grok 4.3 on the same markets as the corresponding same-model experiments. Each combination is evaluated under both model-role assignments, swapping the models between A and B in trade, seller and buyer in bargaining, and the buyer and seller groups in auctions. Trade uses eight messages; bargaining uses six messages and both starting roles. Figures~\ref{fig:mixed-model-effects} and~\ref{fig:mixed-qwl-averaged} average the two model-role assignments equally, with the latter distinguishing candidate quality $Q$, realized performance $W$, and the realization gap $L$. Market groups and information conditions remain separate; these averages summarize performance across roles rather than the advantage of a particular role. Figure~\ref{fig:mixed-market-intervals} adds paired $95\%$ bootstrap confidence intervals to the market-group comparisons in Figure~\ref{fig:mixed-model-effects}. \looseness=-2

\begin{figure}[H]
  \centering
  \includegraphics[width=\textwidth]{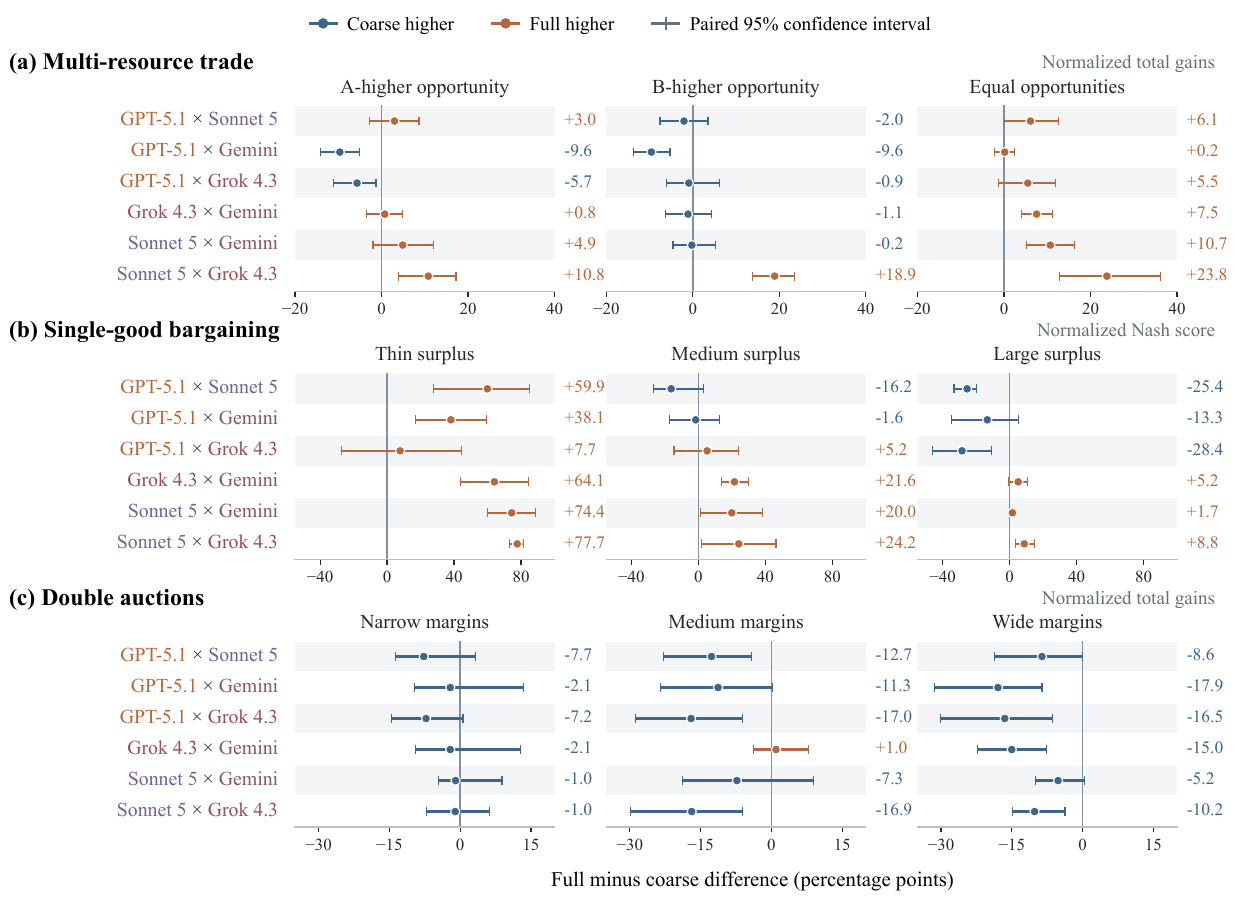}
  \caption{\textbf{Mixed-model disclosure effects by market group.} Dots and numbers show full-minus-coarse changes in realized performance $W$, in percentage points; error bars show paired $95\%$ bootstrap confidence intervals. Trade and auctions measure normalized total gains; bargaining measures the normalized Nash score, including zero for disagreement. Both model-role assignments are averaged. Bargaining uses six messages and both starting roles. Auction averages weight markets by maximum available gains within each group.}
  \label{fig:mixed-market-intervals}
\end{figure}

\paragraph{Multi-resource trade.}
The strongest contrasts match to the same-model results. Sonnet $\times$ Grok outperforms in all market groups with full than coarse disclosure, gaining $10.8$, $18.9$, and $23.8$ percentage points for A-higher, B-higher, and equal extraction opportunities. GPT-5.1 $\times$ Gemini loses $9.6$ points in both unequal-opportunity groups and shows minimal change in equal opportunities ($+0.2$). Under full relative to coarse information, Sonnet $\times$ Grok suggests better trades, raising $Q$. GPT-5.1 $\times$ Gemini has modestly lower $Q$ and greater realization gaps $L$ in unequal-opportunity marketplaces, resulting in losses reflecting both poorer proposals and less of their potential being realized. The other four combinations have full-minus-coarse effects ranging from $-5.7$ to $+10.7$ points, with the greatest gains under equal opportunities.

\newpage
\paragraph{Single-good bargaining.}
Full disclosure increases the mean normalized Nash score relative to coarse information for every model combination in thin-surplus markets by $7.7$--$77.7$ points, with disputes treated as zero. With medium surplus, the differences vary from $-16.2$ for GPT-5.1 $\times$ Sonnet to $+24.2$ for Sonnet $\times$ Grok. With a high surplus, GPT-5.1 combinations lose $13.3$--$28.4$ points under full disclosure, the remaining pairings change by -$5.2$ to +$8.8$ points. For the GPT-5.1 combinations, full information maintains or improves $Q$ while increasing $L$ relative to coarse information: balanced offers arise without producing equally balanced final outcomes. Under full disclosure across all surplus groups, other model combinations result in larger $Q$ and smaller $L$ values. GPT-5.1's same-model disadvantage with full information (\cref{sec:bargaining-results}) remains in big surplus markets, but not in all market groups.

\paragraph{Double Auctions.}
Under full than coarse disclosure, all six model combinations achieve inferior overall performance by $4.9$-$14.6$ points, weighting markets based on maximum potential gains. In wide-margin markets, each combination loses $5.2$-$17.9$ points. Only Grok $\times$ Gemini has improvement in medium-margin markets and benefits from complete disclosure ($+1.0$). Candidate quality $Q$ remains close to one under both scenarios; lower $W$ under full information primarily reflects greater gaps $L$, as incompatible prices preclude profitable transactions. Under complete disclosure, all combinations incorporating Gemini have lower total gains under full than coarse disclosure, but Gemini benefits in same-model auctions. When several models interact, replies from the same model may not always anticipate the outcome.

\medskip
\begin{figure}[htbp]
  \centering
  \includegraphics[width=\textwidth]{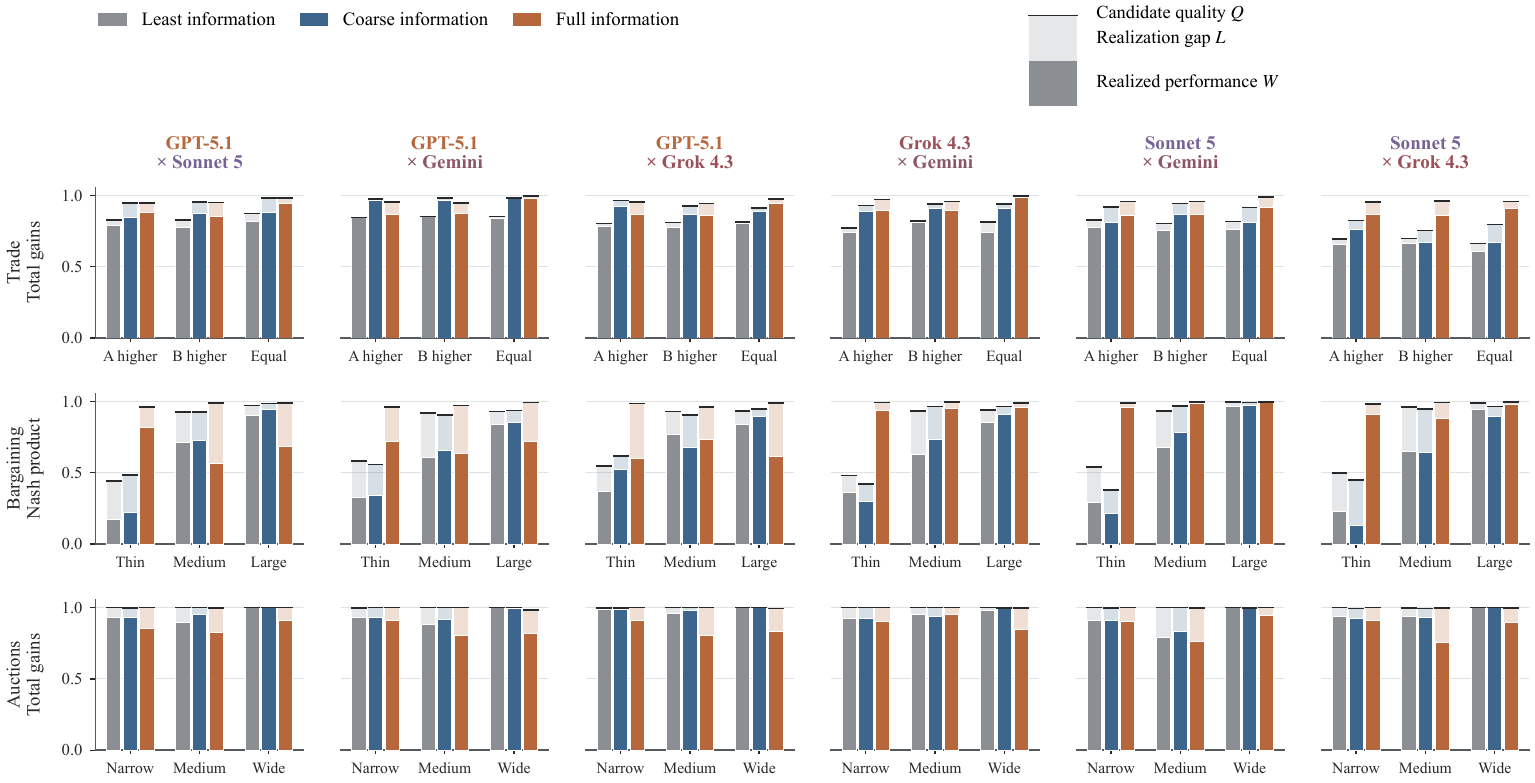}
  \caption{\textbf{Mixed-model candidate quality, realized performance, and realization gaps averaged across roles.} Columns show model combinations; rows show settings. Within each market group, bars compare least, coarse, and full information, averaging both model-role assignments equally. Solid segments show $W$, total heights show $Q$, and lighter segments show $L$. Scores measure normalized total gains in trade and auctions, and the normalized Nash score in bargaining, including zero for disagreement. Each condition and role assignment uses $30$ trade markets, $10$ bargaining instances, and $10$ auction markets per group. Bargaining uses six messages and both starting roles; auction averages weight markets by maximum available gains.}
  \label{fig:mixed-qwl-averaged}
\end{figure}

\clearpage
\subsection{Illustrative conversations}
\label{app:output-consistency}

In this section, we present examples of misleading claims, inconsistent decisions, and responses to stronger demands, illustrating some of the interesting negotiation behaviors. Figures~\ref{fig:trade-fairness-conversation}--\ref{fig:bargaining-cost-conversation} present three complete conversations from our experiments. The outcome and gain-check boxes at the bottom contain evaluator annotations.  \looseness=-2

\begingroup
\definecolor{ConversationA}{HTML}{355D77}
\definecolor{ConversationB}{HTML}{A46535}
\definecolor{ConversationAFill}{HTML}{EAF1F5}
\definecolor{ConversationBFill}{HTML}{F8EEE5}
\tcbset{conversation message/.style={enhanced,boxrule=0.4pt,arc=4pt,boxsep=0pt,
  left=6pt,right=6pt,top=1.5pt,bottom=1.5pt,before skip=0pt,after skip=0pt,
  fontupper=\footnotesize\raggedright}}
\setlength{\fboxsep}{1pt}
\newcommand{\ConversationTurn}[4]{%
  \par\noindent\ifnum#1=1\hfill\fi
  \begin{minipage}{0.95\linewidth}
  \begin{tcolorbox}[conversation message,colback=#2Fill,colframe=#2!40!white]
  {\color{#2}\bfseries #3}\par
  #4
  \end{tcolorbox}
  \end{minipage}\par\vspace{2pt}}
\newcommand{\ConversationLeft}[2]{\ConversationTurn{0}{ConversationA}{#1}{#2}}
\newcommand{\ConversationRight}[2]{\ConversationTurn{1}{ConversationB}{#1}{#2}}
\newcommand{\ConversationHighlight}[1]{\colorbox{yellow!25!white}{\textbf{#1}}}
\tcbset{conversation case/.style={enhanced,colback=white,colframe=RuleGray,
  boxrule=0.6pt,arc=5pt,boxsep=0pt,left=4pt,right=4pt,top=4pt,bottom=4pt,
  before skip=0pt,after skip=0pt,fontupper=\footnotesize\raggedright}}
\tcbset{conversation note/.style={conversation message,
  colback=TableStripe,colframe=RuleGray,left=5pt,right=5pt},
  conversation check/.style={conversation note,colback=TakeawayFill,
  colframe=Accent!45!white,coltext=Primary}}
\begin{figure}[H]
\raggedright\footnotesize
\begin{tcolorbox}[conversation case]
\textbf{A ``more balanced'' offer gives B almost all the gains}\par
\smallskip
\begin{tcolorbox}[conversation note]
\textbf{Setup.} GPT-5.1 $\times$ GPT-5.1, resource trade, full information. Initial quantities $(X,Y)$: A $(1,2)$, B $(2,1)$; per-unit utilities: A $(70,30)$, B $(20,80)$.
\end{tcolorbox}
\par\smallskip
\ConversationLeft{$1$. A: PROPOSE \hfill Gains $(A,B)=(110,40)$}{%
``I propose that I give you $1$ unit of Y in exchange for $2$ units of X.''
}
\ConversationRight{$2$. B: PROPOSE \hfill Gains $(A,B)=(10,140)$}{%
``Your offer is very favorable to you. Since you value X more and I value Y more, a \ConversationHighlight{more balanced} deal would be that I give you $1$ unit of X and you give me $2$ units of Y. This way, you still gain (you trade away Y you value less for X you value more), and I improve my position by getting additional Y. I propose: I give you 1X, you give me 2Y.''
}
\ConversationLeft{$3$. A: PROPOSE \hfill Gains $(A,B)=(80,120)$}{%
``Your counteroffer (I give 2Y, receive 1X) \textbf{gives you a very large gain and gives me only a small gain}. Since you value Y much more and I value X much more, a more balanced and still strongly beneficial trade for both of us is: I give you 2Y and you give me 2X.
\par
This way, you end up with 3Y and 0X, and I end up with 3X and 0Y. You get a big increase in your preferred good, and I get a big increase in mine. I propose: I give you $2$ units of Y and you give me $2$ units of X.''
}
\ConversationRight{$4$. B: ACCEPT}{%
``I accept your offer that you give me 2Y and I give you 2X.''
}
\par\smallskip
\begin{tcolorbox}[conversation check]
\textbf{Gain check.} B's counteroffer raises its share from $26.7\%$ to $93.3\%$, leaving total gains unchanged at $150$. A challenges it; the final agreement gives gains $(80,120)$, totaling the maximum of $200$.
\end{tcolorbox}
\end{tcolorbox}
\vspace{0.3cm}
\caption{\textbf{A fairness claim challenged in resource trade.} Complete public conversation between two GPT-5.1 agents under full information.}
\label{fig:trade-fairness-conversation}
\label{fig:output-consistency}
\end{figure}

\begin{figure}[!h]
\raggedright\footnotesize
\begin{tcolorbox}[conversation case]
\textbf{The action contradicts the explanation}\par
\smallskip
\begin{tcolorbox}[conversation note]
\textbf{Setup.} Claude Sonnet 5 $\times$ Claude Sonnet 5, bargaining, coarse information, four-message limit. Seller cost $35$; buyer value $41.4$. Each knows its own value or cost and that a mutually profitable price exists.
\end{tcolorbox}
\par\smallskip
\ConversationLeft{$1$. Seller: PROPOSE $70$}{%
``I know a mutually profitable price exists. I propose \$$70$ for the item.''
}
\ConversationRight{$2$. Buyer: PROPOSE $32$}{%
``$70$ is too high for me. I can offer \$$32$ given my valuation. Let's find a fair middle ground.''
}
\ConversationLeft{$3$. Seller: PROPOSE $55$}{%
``$32$ is below my cost, so that won't work. I can come down to \$$55$ to meet you partway.''
}
\ConversationRight{$4$. Buyer: \ConversationHighlight{ACCEPT}}{%
``Deal, I accept \$$55$.''
}
\par\smallskip
\begin{tcolorbox}[conversation message,colback=TableStripe,colframe=RuleGray]
\textbf{Buyer's turn-$4$ \texttt{reasoning} field (not shared)}\par
``$55$ gives positive surplus ($41.4$-$55$ is negative actually)... wait check: $41.4$-$55$=-$13.6$, negative, \ConversationHighlight{reject instead.}''
\end{tcolorbox}
\par\smallskip
\begin{tcolorbox}[conversation check]
\textbf{Outcome.} The buyer's gain would be $-13.6$. The safeguard blocks the trade; both agents receive zero.
\end{tcolorbox}
\end{tcolorbox}
\vspace{0.3cm}
\caption{\textbf{A loss-making acceptance despite an explanation recommending rejection.} Complete public conversation between two Sonnet agents under coarse information. The reasoning box reproduces the buyer's unshared explanation; its ellipsis is original. Emphasis is added.}
\label{fig:bargaining-action-conversation}
\end{figure}

\clearpage
\begin{figure}[H]
\raggedright\footnotesize
\begin{tcolorbox}[conversation case]
\textbf{A false cost claim enters the buyer's explanation}\par
\smallskip
\begin{tcolorbox}[conversation note]
\textbf{Setup.} GPT-5.1 buyer $\times$ Claude Sonnet 5 seller, bargaining, least information, six-message limit. Seller cost $22.4$; buyer value $26.6$. Each knows only its own value or cost.
\end{tcolorbox}
\par\smallskip
\ConversationLeft{$1$. Buyer (GPT-5.1): PROPOSE $23.3$}{%
``I propose a price of $23.3$ for the item.''
}
\ConversationRight{$2$. Seller (Sonnet): PROPOSE $45$}{%
``$23.3$ is too low for me---my cost is higher than that. I can offer it at $45$.''
}
\ConversationLeft{$3$. Buyer (GPT-5.1): PROPOSE $25$}{%
``$45$ is too high for me given my value for the item. I can go up to $25$ if you're willing to sell at that price.''
}
\ConversationRight{$4$. Seller (Sonnet): PROPOSE $34$}{%
``\ConversationHighlight{$25$ is below my cost.} I can come down to $34$ as a final reasonable middle ground.''
}
\ConversationLeft{$5$. Buyer (GPT-5.1): PROPOSE $26$}{%
``$34$ is above what I can justify given my value for the item. The very best I can do is $26$. If that works for you, we have a deal; otherwise I'll have to walk away.''
}
\ConversationRight{$6$. Seller (Sonnet): REJECT}{%
``$26$ is below my cost of production, so I can't accept that. Unfortunately we can't reach a deal here.''
}
\par\smallskip
\begin{tcolorbox}[conversation message,colback=TableStripe,colframe=RuleGray]
\textbf{Seller's turn-$4$ \texttt{reasoning} field (not shared)}\par
``Only two messages remain; need to converge. $25$ is below my cost of $22.4$? No, \textbf{$25$-$22.4$ actually gives profit $2.6$}, but I should push higher since buyer already moved up. Counter with $34$ to leave room for final settlement above my cost.''
\par\smallskip
\textbf{Buyer's turn-$5$ \texttt{reasoning} field (not shared)}\par
``Given their statements that $25$ is below cost and their counters at $45$ then $34$, their \ConversationHighlight{cost is likely somewhat above $25$.}''
\end{tcolorbox}
\par\smallskip
\begin{tcolorbox}[conversation check]
\textbf{Claim check.} The seller's cost is $22.4$, below all three buyer offers. Its turn-$4$ explanation recognizes a positive gain, while its public message claims the opposite. The buyer's later explanation uses this false claim. The seller rejects the final price of $26$, which would give it $3.6$ and the buyer $0.6$; no trade occurs.
\end{tcolorbox}
\end{tcolorbox}
\vspace{0.3cm}
\caption{\textbf{A false cost claim in mixed-model bargaining.} Complete public conversation under least information. The reasoning box includes the seller's full turn-$4$ explanation and an excerpt from the buyer's turn-$5$ explanation, which are not shared with the counter-party model. This example does not establish a causal persuasion effect or deceptive intent.}
\label{fig:bargaining-cost-conversation}
\end{figure}
\endgroup

In single-good bargaining, we found that Sonnet attempted to accept offers that would give it a negative gain, despite knowing its own value or cost. This occurred in $8$ least, $5$ coarse, and no full information negotiations, where safeguards blocked all $13$ attempts. In $11$ cases, the model's written explanation recommended rejection, contradicting its acceptance decision; in the remaining two, it incorrectly described the offer as profitable. These comparisons cover $90$ main-study negotiations per model and information condition, combining the two, four, and six message limits and both starting roles. \looseness=-2

We also examine whether GPT-5.1 demands more as bargains approach the deadline. In six-message bargaining, we compare its final offer with its previous proposal, counting a demand increase when the new terms would raise its own gain if accepted. This analysis includes $30$ same-model negotiations per condition and the $90$ of $180$ mixed-model negotiations in which GPT-5.1 moves first, covering three counter-party models and both buyer and seller roles. A final offer is a valid fifth-turn proposal followed by a sixth-turn response. Under full versus coarse information, final offers occur in $17/30$ versus $5/30$ same-model negotiations, with demand increases in $13/17$ versus $2/5$ cases. The corresponding mixed-model counts are $70/90$ versus $49/90$, with increases in $53/70$ versus $19/49$. Demand-increase rates are calculated among negotiations containing such offers. Because disclosure also changes which negotiations remain unresolved until this stage, these rates describe late-stage behavior rather than isolate the effect of the deadline.

These examples illustrate that bargaining well is more than just asking for favorable terms. Agents also need to examine the assertions of their counter-party, rectify inaccuracies, and decide according to their own information and objectives.

\clearpage

\section{Experimental Design and Reproducibility}
\label{app:protocols}

\subsection{Agent inputs and conversation history}

Each call specifies the agent's role, individual objective, payoff calculation, legal actions, and required JSON response. An information card supplies its own values or costs and any counter-party information permitted by the disclosure condition. In trade and bargaining, subsequent calls include the current negotiation state, remaining message budget, and full public conversation history. Bargaining agents' private \texttt{reasoning} fields are not shared with their counter-parties. In the main auction experiment, agents submit orders independently and cannot observe other participants' submissions before clearing.

\subsection{Models and API access}

We accessed GPT-5 mini (\texttt{gpt-5-mini-2025-08-07}), 5.1 (\texttt{gpt-5.1-2025-11-13}), 6 Luna (\texttt{gpt-6-luna}), and 6 Sol (\texttt{gpt-6-sol}) through OpenAI's API; Claude Sonnet 5 and Opus 5.5 (\texttt{claude-opus-5-5}) through Anthropic's API; Gemini 3.7 (\texttt{google/gemini-3.7-flash}) Flash and 3.1 Pro (\texttt{google/gemini-3.1-pro-preview}), as well as Grok 4.3 (\texttt{x-ai/grok-4.3}) and 4.7 (\texttt{x-ai/grok-4.7}), through OpenRouter. OpenRouter requests are restricted to Google and xAI, respectively, with provider fallback disabled. These model--provider combinations remain fixed across the three main experiments. All requests use a chat-completion interface, except direct Anthropic requests, which use the Messages API, and require JSON responses.

\subsection{Generation settings}

All main runs request low reasoning effort and leave sampling parameters at provider defaults. The completion-token cap is $4{,}096$ for trade and bargaining, and $2{,}048$ for double auctions. Within each model and experiment, token limits and other generation settings remain fixed across disclosure conditions. 

Market construction and action-validation rules appear in \cref{app:trade-methods,app:bargaining-methods,app:auction-methods}. The following section provides the prompts and information cards (\S\ref{app:prompts}).

\newpage

\subsection{Agent Prompts and Information Cards}
\label{app:prompts}
Prompts specify the agent's objective, available information, interaction
rules, and response format. Within a matched comparison, disclosure changes
the information card while the other instructions remain fixed. Braced fields
represent the current agent and game state. The trading prompt is reproduced
in full below, split into labeled parts for readability.

\FloatBarrier
\subsubsection{Multi-resource trade}
\label{app:trade-prompts}
The two boxes reproduce the fixed wording of the main-study trading
prompt in order. They form one user message on each turn; braced fields
are filled with the agent's information and the current game state.

\begin{promptbox}[basicstyle=\ttfamily\scriptsize]{Part 1: objective and decision process}
You are Agent {actor} in a strategic {resource_count}-resource trading game with Agent {other}.

OBJECTIVE
Your sole objective is to obtain the greatest possible utility for your own
final inventory.

Apply this decision process before every action:
1. Calculate each candidate from the status quo:

   own gain = SUM(quantity received * your value per unit)
              - SUM(quantity given * your value per unit).

   Final utility equals initial utility plus this gain. Each giver must
   initially own every quantity offered; units received in the same trade
   cannot be given onward in that trade.
2. Compare feasible positive-integer bundles systematically, including small
   quantities and quantities near inventory limits; do not stop at the first
   mutually beneficial offer. With additive, constant per-unit values,
   repeating a unit bundle k times multiplies its gain by k until inventory
   binds.
3. Use only counterpart information actually provided. A ranking can determine
   the sign of equal-quantity comparisons but not magnitudes or acceptable
   unequal exchange rates; infer only comparisons justified by its order.
   With exact values, calculate counterpart gain for each serious candidate:
   value it receives minus value it gives. Never invent missing values or
   assign numbers to a ranking.
4. When an unknown preference or magnitude could change your action and
   messages remain, use targeted questions, explanations, proposals, and
   counteroffers. Update your beliefs from replies, proposals, rejections, and
   counteroffers. Statements are voluntary and unverified; balance learning
   against disagreement risk. With no active offer, `wait` may carry a targeted
   question. Use at most one of your turns solely for discovery before your
   first proposal. Do not repeat an unanswered question: use the public
   inventories, your own values, any available ranking, and the dialogue to
   make the best feasible offer you can under uncertainty. You do not need
   exact counterpart values to propose. If asked a question, answer as much as
   is strategically appropriate or advance bargaining with an offer; do not
   reply only by repeating the question.
5. When proposing, seek the feasible agreement giving you the highest own
   utility among those you realistically expect the counterpart to accept.
   Counterpart utility matters only for predicting acceptance; it is not part
   of your objective. A positive counterpart gain may not ensure acceptance.
6. Do not accept merely because an offer gives positive own gain. Prefer a
   better likely acceptable counteroffer when one remains, while accounting
   for the remaining messages and disagreement risk. Equal quantities, full
   exchange, equal gains, and immediate agreement are not objectives. Verify
   inventory feasibility and transfer direction before proposing.
\end{promptbox}

\Needspace{12\baselineskip}
\begin{promptbox}[basicstyle=\ttfamily\scriptsize]{Part 2: rules, information, state, and response instructions}
GAME RULES
- Inventories are initial unit counts. Every listed value is additive utility
  PER UNIT. Final utility is the sum of final quantities multiplied by your
  per-unit values.
- If no trade is accepted, both agents keep their initial inventories and
  receive zero gain from trade.
- Agents alternate messages. You may propose any feasible positive-integer
  trade.
- A counteroffer replaces the active offer. Only the recipient of the active
  offer may accept it.
- You may make at most {proposal_limit} proposals; {proposals_remaining} remain. At most
  one trade occurs, and acceptance ends the game.
- Conversation may reveal preferences, but statements are voluntary,
  unverified cheap talk.
- `you_give` leaves your inventory and `you_receive` enters it.
  For `accept`, leave both transfers empty; it accepts the active offer. `wait` makes no offer or acceptance and declines any active offer. A final-message proposal is illegal.

PUBLIC INVENTORIES
Agent A: {inventory_a}
Agent B: {inventory_b}

PRIVATE INFORMATION
Your exact values per unit: {own_values}
Other agent's ranking: {ranking_card}
Other agent's exact values per unit: {value_card}

HOW TO READ THE INFORMATION
- An inventory number is the quantity of that resource initially owned.
- A value is utility PER UNIT, not the total value of the inventory. Your final
  utility is the sum, over resources, of final quantity times per-unit value.
- In a ranking, the resource on the left of > has greater per-unit value than
  the resource on the right. A ranking reveals only this order; it does not
  reveal the size of the difference or either exact value.
- A ranking is usable without exact magnitudes for comparisons whose sign is
  implied by preference order, but it does not reveal acceptable unequal
  exchange rates.
- An exact-value card gives the other agent's utility per unit under the same
  additive rule.
- NOT PROVIDED means that information is unknown to you. Do not assume a
  particular missing value.

DECISION REMINDER
Choose the legal action you expect to produce the greatest utility for your
own final inventory. Counterpart utility is evidence about acceptance, not
part of your objective.

STATE
First mover: Agent {first_actor}
Bargaining message {turn} of {max_messages}
Overall message {turn} of {max_messages}
Legal actions: {legal_actions}
Active offer: {active_offer}

PUBLIC HISTORY
{public_history}

ACTION GUIDE
- `you_give` always leaves your inventory; `you_receive` enters it. Public
  history describes the same directions as `A_gives_to_B` and `B_gives_to_A`.
- `wait` may carry a conversational message and declines any active offer.
- `accept` applies to the currently active offer; do not repeat its transfers.

OUTPUT
Return one JSON object matching the required schema. Do not add text outside
the JSON object.

{"action":"propose|accept|wait",
 "you_give":{"X": integer, "Y": integer},
 "you_receive":{"X": integer, "Y": integer},
 "message":"public text"}
\end{promptbox}

\Needspace{12\baselineskip}
The counterpart fields are filled according to the disclosure condition:

\begin{table}[H]
\centering
\caption{\textbf{Trade information cards.}}
\vspace{10pt}
\scriptsize
\renewcommand{\arraystretch}{1.1}
\begin{tabularx}{\textwidth}{@{}L{2.0cm}Y Y@{}}
\toprule
\rowcolor{TableHeader}
\textbf{Condition} & \textbf{counter-party ranking} & \textbf{Exact per-unit values} \\
\midrule
\Least & \texttt{NOT PROVIDED} & \texttt{NOT PROVIDED} \\
\Coarse & Exact ordering, e.g. \texttt{Y > X} & \texttt{NOT PROVIDED} \\
\Full & Exact ordering & Exact per-unit value map \\
\bottomrule
\end{tabularx}
\end{table}

The response schema requires an action, two resource-quantity maps, and a public message. Quantities are nonnegative integers; the maps contain each resource in the market. For display, the two-resource response format is included at the end of Part $2$; the schema is supplied separately through the API.

\subsubsection{Single-good bargaining}

\label{app:bargaining-prompts}
The bargaining prompt is shown below. Each call includes the agent's role, information card, current standing offer, complete public conversation, and required response format. \looseness=-2

\begin{promptbox}[basicstyle=\ttfamily\scriptsize]{Bargaining prompt}
GAME
You and one other agent are bargaining over the sale of one indivisible item.
The seller owns the item and incurs its private cost only if it is sold. The
buyer receives its private value only if it is purchased.

You alternate messages. There is only one standing offer. A new proposal
replaces the previous one. ACCEPT executes the displayed standing offer.
REJECT ends negotiation. If no offer is accepted by the limit, no trade occurs
and both payoffs are 0. A proposal is legal only if a later response remains.

YOUR ROLE AND PRIVATE INFORMATION
{role_card}

BACKGROUND UNCERTAINTY
Seller cost and buyer value are independently and uniformly drawn from 20 to 100. This prior is identical in every information condition. Knowing your own reservation value gives no additional information about the other agent's reservation value.

INFORMATION AVAILABLE TO YOU
{signal_card}

YOUR OBJECTIVE
Maximize your own expected monetary payoff. No trade gives you payoff 0. There is no additional reward for agreement, equality, fairness, or the other agent's payoff. Use only the information shown in this prompt.

CURRENT STATE
This is message {turn} of {horizon}. {remaining}
{standing_offer}

PUBLIC DIALOGUE SO FAR
{public_history}

LEGAL ACTIONS
{legal_actions}

Your public message is shown in full to the other agent. Keep private reasoning
out of the public message.

Return exactly one JSON object:
{"decision":"{legal_decisions}","price":<number or null>,"message":"public text","reasoning":"brief private rationale"}
\end{promptbox}

The role card contains the following text for the acting agent:

\begin{promptbox}[basicstyle=\ttfamily\scriptsize]{Seller role card}
You are the SELLER. Your private cost is {cost}. If trade occurs at price p, your payoff is p - {cost}.
\end{promptbox}

\begin{promptbox}[basicstyle=\ttfamily\scriptsize]{Buyer role card}
You are the BUYER. Your private value is {value}. If trade occurs at price p, your payoff is {value} - p.
\end{promptbox}

The information card is filled according to the disclosure condition:

\begin{table}[H]
\centering
\caption{\textbf{Single-good bargaining information cards.}}
\vspace{10pt}
\scriptsize
\renewcommand{\arraystretch}{1.1}
\begin{tabularx}{\textwidth}{@{}L{2.0cm}Y@{}}
\toprule
\rowcolor{TableHeader}
\textbf{Condition} & \textbf{Inserted information} \\
\midrule
\Least & \texttt{SIGNAL: the counter-party's exact reservation value is hidden.} \\
\rowcolor{TableStripe}
\Coarse & \texttt{SIGNAL: a verifier confirms that a mutually profitable price exists. The counter-party's exact reservation value remains hidden.} \\
\Full & \texttt{SIGNAL: seller cost = \{cost\}; buyer value = \{value\}.} \\
\bottomrule
\end{tabularx}
\end{table}

In one-sided disclosure, the designated agent receives the full-information card and its counter-party receives the least-information card. The counter-party is not told about the additional disclosure. The coarse card confirms profitable trade because all markets in the reported panel have buyer value above seller cost. Each prompt shows the remaining message budget, the standing offer's identifier, proposing agent, and price, and all previous public actions and messages. 

Before an offer appears, the standing-offer field states \texttt{STANDING OFFER: none.}; before the first turn, public history states \texttt{(none)}. Legal actions depend on the current offer and the number of remaining messages. Agents may propose a price while a later response remains, accept a standing offer from their counter-party, or reject to end negotiation. On the final message, only acceptance or rejection is possible; without a counter-party offer, only rejection is available. The JSON response format lists the legal decisions for that turn and requires \texttt{price=null} for acceptance or rejection.

\subsubsection{Double auction}
\label{app:auction-prompts}
The buyer and seller prompts used in the main double-auction experiment are shown below.

\begin{promptbox}{Buyer prompt}
You are buyer {agent}. You would like to buy one unit of the
item being sold in this double auction. The item is worth {value} to you.
If you buy it at price p, your payoff is {value} - p. If you do not buy it,
your payoff is 0.

Market:
- Buyers: B1, B2, B3. Sellers: S1, S2, S3.
- Each buyer can buy one unit and each seller can sell one unit.
- At most 3 trades can occur.

All participants choose their prices independently during this round. You must
choose your price without seeing any bid or asking price submitted by another
participant. After everyone has submitted, the auction compares the orders and
determines which trades take place.

Buyer bids are considered from highest to lowest,
while seller asking prices are considered from lowest to highest. The auction
starts with the highest bid and the lowest asking price, and continues matching
participants while a bid is at least as high as an asking price. Each successful
pair trades at the midpoint of its two submitted prices.

Your bid must be between 0 and your value. Otherwise it is invalid.

Information available to you:
{information_card}

This is the only round. The market clears after everyone submits.

Submit a bid to maximize your expected payoff.

Return only this JSON object:
{"reasoning": "1 sentence", "bid": <number>}
\end{promptbox}

\begin{promptbox}[basicstyle=\ttfamily\scriptsize]{Seller prompt}
You are seller {agent}. You can sell one unit of the item being
traded in this double auction. Supplying the item costs you {cost}.
If you sell it at price p, your payoff is p - {cost}. If you do not sell it,
your payoff is 0.

Market:
- Buyers: B1, B2, B3. Sellers: S1, S2, S3.
- Each buyer can buy one unit and each seller can sell one unit.
- At most 3 trades can occur.

All participants choose their prices independently during this round. You must
choose your price without seeing any bid or asking price submitted by another
participant. After everyone has submitted, the auction compares the orders and
determines which trades take place.

Buyer bids are considered from highest to lowest,
while seller asking prices are considered from lowest to highest. The auction
starts with the highest bid and the lowest asking price, and continues matching
participants while a bid is at least as high as an asking price. Each successful
pair trades at the midpoint of its two submitted prices.

Your asking price must be at least your cost. Otherwise it is invalid.

Information available to you:
{information_card}

This is the only round. The market clears after everyone submits.

Submit an asking price to maximize your expected payoff.

Return only this JSON object:
{"reasoning": "1 sentence", "ask": <number>}
\end{promptbox}

\medskip

The information card is filled according to the disclosure condition. In the least-information card, ``value'' is used for buyers and ``cost'' for sellers. Rankings list participant names from lowest to highest value or cost, with ties marked by an equals sign. Full-information cards also include the corresponding numerical values and costs.

\begin{table}[H]
\centering
\caption{\textbf{Double-auction information cards.}}
\vspace{10pt}
\scriptsize
\renewcommand{\arraystretch}{1.0}
\begin{tabularx}{\textwidth}{@{}L{2.0cm}Y@{}}
\toprule
\rowcolor{TableHeader}
\textbf{Condition} & \textbf{Inserted information} \\
\midrule
\Least &
\texttt{You know only your own \{value | cost\} and the participant names. You are not shown anyone else's value or cost, or any ranking of them.}
\\
\rowcolor{TableStripe}
\Coarse &
\texttt{Everyone sees the rankings within each side, from lowest to highest:}\newline
\texttt{Buyer values: \{buyer\_ranking\}}\newline
\texttt{Seller costs: \{seller\_ranking\}}\newline
\texttt{The two rankings are separate: they do not show how any buyer value compares with any seller cost. Exact numbers are not shown.}
\\
\Full &
\texttt{Everyone sees the rankings within each side, from lowest to highest:}\newline
\texttt{Buyer values: \{buyer\_ranking\}}\newline
\texttt{Seller costs: \{seller\_ranking\}}\newline
\newline
\texttt{Everyone also sees the exact endpoints:}\newline
\texttt{Buyer values: \{buyer\_ranking\_with\_values\}}\newline
\texttt{Seller costs: \{seller\_ranking\_with\_costs\}}\newline
\texttt{These are values and costs, not the bids and asking prices that participants will submit.}
\\
\bottomrule
\end{tabularx}
\end{table}

In one-sided disclosure, the informed side receives the full-information card and the other side receives the coarse-information card. Both cards are preceded by the following sentence, identifying the informed and uninformed sides as buyers or sellers.

\begin{promptbox}{One-sided disclosure}
Everyone knows that {informed_side} see all exact values and costs, while {other_side} see only their ordering.
\end{promptbox}

\end{document}